\documentclass[manuscript, screen]{acmart}
\usepackage{array}
\usepackage{multirow}
\usepackage{xcolor}
\AtBeginDocument{%
  }

\begin{document}

%%
%% The "title" command has an optional parameter,
%% allowing the author to define a "short title" to be used in page headers.
\title{Beyond the Norm: A Survey of Synthetic Data Generation for Rare Events}

%%
%% The "author" command and its associated commands are used to define
%% the authors and their affiliations.
%% Of note is the shared affiliation of the first two authors, and the
%% "authornote" and "authornotemark" commands
%% used to denote shared contribution to the research.
\author{Jingyi Gu}
\authornote{Both authors contributed equally to this research.}
\email{jg95@njit.edu}
\orcid{0000-0001-8010-8326}
\author{Xuan Zhang}
\authornotemark[1]
\email{xz296@njit.edu}
\orcid{0009-0006-7251-3077}
\author{Guiling Wang}
\email{gwang@njit.edu}
\orcid{0000-0003-1880-4763}
\affiliation{%
  \institution{New Jersey Institute of Technology, Newark, NJ}
  \country{USA}
}

%%
%% By default, the full list of authors will be used in the page
%% headers. Often, this list is too long, and will overlap
%% other information printed in the page headers. This command allows
%% the author to define a more concise list
%% of authors' names for this purpose.
% \renewcommand{\shortauthors}{Trovato et al.}

%%
%% The abstract is a short summary of the work to be presented in the
%% article.

\begin{abstract}

Extreme events, such as market crashes, natural disasters, and pandemics, are rare but catastrophic, often triggering cascading failures across interconnected systems. Accurate prediction and early warning can help minimize losses and improve preparedness. While data-driven methods offer powerful capabilities for extreme event modeling, they require abundant training data—yet extreme event data is inherently scarce, creating a fundamental challenge. Synthetic data generation has emerged as a powerful solution. However, existing surveys focus on general data with privacy preservation emphasis, rather than extreme events' unique performance requirements. This survey provides the first overview of synthetic data generation for extreme events. We systematically review generative modeling techniques and large language models, particularly those enhanced by statistical theory as well as specialized training and sampling mechanisms to capture heavy-tailed distributions. We summarize benchmark datasets and introduce a tailored evaluation framework covering statistical, dependence, visual, and task-oriented metrics. A central contribution is our in-depth analysis of each metric's applicability in extremeness and domain-specific adaptations, providing actionable guidance for model evaluation in extreme settings. We categorize key application domains and identify underexplored areas like behavioral finance, wildfires, earthquakes, windstorms, and infectious outbreaks. Finally, we outline open challenges, providing a structured foundation for advancing synthetic rare-event research.

\end{abstract}

%%
%% The code below is generated by the tool at http://dl.acm.org/ccs.cfm.
%% Please copy and paste the code instead of the example below.

%%
%% Keywords. The author(s) should pick words that accurately describe
%% the work being presented. Separate the keywords with commas.

% \received{20 February 2007}
% \received[revised]{12 March 2009}
% \received[accepted]{5 June 2009}

%%
%% This command processes the author and affiliation and title
%% information and builds the first part of the formatted document.
\maketitle

\section{Introduction}
Extreme events—rare, high-impact occurrences that deviate significantly from normal expectations—pose severe consequences across domains, yet have received limited attention.
Black Swan events such as the 1987 Black Monday crash, the 2008 global financial crisis, and the 2010 Flash Crash erased trillions in market value within hours or days. Natural disasters like Hurricane Katrina in 2005 caused massive infrastructure collapse and long-term population displacement, while the COVID-19 pandemic triggered unprecedented global disruption across healthcare systems, supply chains, and human behavior.
Given their disproportionately large societal and economic impacts, the early detection and timely warning of extreme events have become crucial priorities for improving preparedness, maintaining societal resilience, and minimizing catastrophic losses.

Data-driven methods and artificial intelligence (AI) techniques offer potentially powerful capabilities to detect anomalous patterns and predict extreme events, such as financial crashes or natural disasters, helping institutions quantify and prepare for extreme scenarios. However, these approaches fundamentally depend on abundant training data—a requirement that extreme events cannot satisfy due to their inherent rarity compared to normal conditions.
The insufficiency of high-quality extreme data makes it particularly difficult for models to train efficiently and capture the unique abnormal patterns, complex dependencies, and cascading behaviors that characterize high-impact occurrences. This limitation poses significant challenges across applications, including financial risk forecasting, climate resilience planning, epidemic response, and infrastructure protection.

Synthetic data generation has emerged as a promising solution to bridge this critical gap. By generating artificial extreme event data to enrich sparse datasets, synthetic data can provide machine learning models with the abundant training examples necessary for effective learning. Generative models, including generative adversarial networks (GANs), variational autoencoders (VAEs), and diffusion models, are commonly employed to simulate extreme scenarios and augment sparse datasets with plausible instances of extreme conditions. This approach transforms data-scarce modeling challenges into data-sufficient opportunities for robust algorithm development. 

While numerous surveys on synthetic data generation across various domains \cite{figueira2022survey,assefa2020generating,gonzales2023synthetic,meiser2024survey,pulla2024synthetic}, none have specifically focused on extreme events. This gap is particularly significant because synthetic data generation for extreme events differs fundamentally from conventional approaches. 
Traditional synthetic data generation emphasizes privacy preservation because general datasets often contain personal or confidential information that cannot be directly shared due to regulatory constraints and ethical concerns, making synthetic alternatives necessary for data sharing and collaboration. 
However, extreme data synthesis focuses primarily on extremeness performance: whether generated samples can effectively capture the rare, high-impact characteristics that define crisis scenarios. 
Extreme event synthesis must model statistical anomalies, temporal clustering behaviors, and cascading failure dynamics—challenges absent in conventional approaches. Consequently, this field requires novel algorithms for heavy-tailed distributions, benchmark datasets for extreme scenarios, and evaluation metrics that assess crisis-relevant performance rather than general similarity.

\begin{figure*}[h]
    \centering
\includegraphics[width=\linewidth]{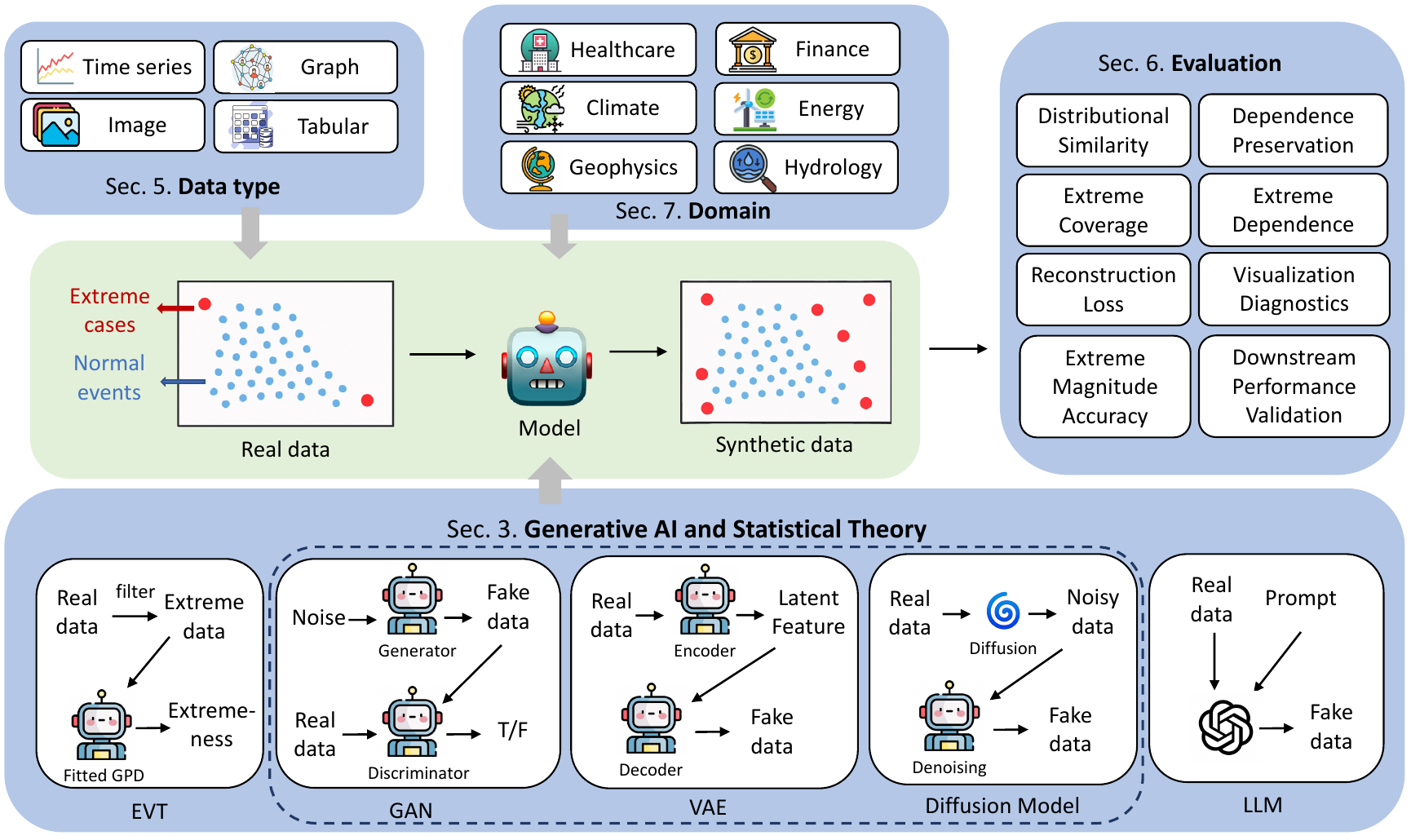}
    % \vspace{2mm}
    \caption{The overall framework of our survey.}
    \Description{A conceptual diagram illustrating the generation of extreme data using generative models and statistical theory. }
    \label{fig:framework}
\end{figure*}

This survey presents the first comprehensive overview of synthetic data generation task tailored for extreme events, as shown in Figure \ref{fig:framework}.
We begin with technical background and methods, systematically review generative models—enhanced by statistical theory and specialized training and sampling mechanisms—as well as large language models (LLMs) to simulate complex tail behaviors. 
We then examine benchmark datasets commonly used for existing studies, followed by a central contribution of this work: a comprehensive evaluation framework tailored for extreme event modeling, moving beyond prior surveys with privacy focus. This framework offers in-depth insights from each metric's applicability in assessing rare-event behavior to its domain-specific adaptions, including advantages and limitations. It organizes metrics across critical dimensions, from distributional similarity and extreme coverage to visualization and task-based validation. 
Next, we categorize the key applications affected by extreme events, such as finance, healthcare, climate, hydrology, energy, and geophysics, while identifying underexplored yet high-stake areas, such as behavioral finance, widespread infectious outbreaks, wildfires, and earthquakes. 
Finally, we outline current challenges and future research opportunities based on identified gaps in the literature. Our goal is to provide a structured foundation for researchers and practitioners while inspiring further innovations in this area.
Our contributions are summarized as follows:
\begin{enumerate}
    \item \textbf{First Comprehensive Survey of Synthetic Extreme Data Generation}: To the best of our knowledge, this is the first survey that systematically covers all major aspects of synthetic extreme data research, including generative models, representative datasets, application domains, and evaluation metrics. 
    \item \textbf{Structured Evaluation Framework with In-Depth Insights}: We introduce a structured evaluation framework tailored to extreme event modeling, moving beyond conventional synthetic data assessment that emphasizes privacy preservation. The framework provides detailed theoretical definitions and in-depth insights into each metric's applicability in extremes and domain-specific adaptations. It organizes metrics into eight categories, covering statistical similarity, dependence preservation, extreme coverage, visualization quality, and downstream task performance, enabling rigorous assessment of rare-event realism and utility.
    \item \textbf{Domain Expansion and Research Outlook}: We highlight underexplored but high-impact domains such as behavioral finance, infectious outbreaks, natural disasters, and outline key research challenges across technical and application domains. These insights aim to inform future directions and spark continued innovation in synthetic extreme data generation.
\end{enumerate}

The remainder of this paper is organized as follows: Section \ref{sec:overview} provides the problem formulation for synthetic data generation for extreme data. Section \ref{sec:tech} introduces technical background on extreme value theory and generative models. Section \ref{sec:method} reviews modeling methods including generative models enhanced by traditional statistical approaches and LLMs. Section \ref{sec:data} presents benchmark datasets. Section \ref{sec:metric} organizes evaluation metrics for extremeness performance assessment. Section \ref{sec:domain} categorizes application domains and identifies underexplored areas. Section \ref{sec:challenge} outlines current research challenges and future opportunities. Finally, Section \ref{sec:conclusion} concludes the survey.

\section{Problem Formulation}\label{sec:overview}

Figure 1 illustrates the comprehensive framework for synthetic extreme data generation. The process begins with real data containing both normal events and rare extreme cases, which are characterized by their scarcity and high-impact nature. This imbalanced dataset is fed into generative models that learn to capture the complex patterns and statistical properties of extreme events. The core challenge lies in training models to effectively distinguish and reproduce the rare extreme patterns while not being overwhelmed by the abundant normal cases. Once trained, these models generate synthetic data that augments the original sparse extreme event dataset, transforming data-scarce modeling challenges into data-sufficient opportunities. The framework encompasses multiple data types (time series, graph, image, tabular) and spans various application domains (healthcare, finance, climate, energy, geophysics, hydrology). The quality and effectiveness of the generated synthetic data are assessed through a comprehensive evaluation framework that examines distributional similarity, dependence preservation, extreme coverage, dimensionality reduction visualization, reconstruction quality, extreme dependence, and downstream task performance. This systematic approach enables robust extreme event modeling and prediction across diverse high-stakes applications where traditional data-driven methods fall short due to insufficient extreme event samples.
\section{Technical Background}\label{sec:tech}

\subsection{Extreme Value Theory (EVT) and Generalized Pareto Distribution (GPD)}
Extreme Value Theory (EVT) \cite{haan2006extreme} provides the theoretical foundation for modeling the statistical behavior of rare, extreme events. It focuses on the asymptotic distribution of maxima or threshold exceedances in a sequence of random variables. EVT consists of two main approaches: the Block Maxima method \cite{ferreira2015block}, which leads to the Generalized Extreme Value (GEV) distribution, and the Peaks Over Threshold (POT) method \cite{leadbetter1991basis}, which results in the Generalized Pareto Distribution (GPD)\cite{castillo1997fitting} for modeling exceedances over a high threshold. The POT framework is particularly useful in continuous settings where rare events occur sporadically, offering more efficient use of data compared to block-based methods.
The cumulative distribution function (CDF) of the GPD is given by:
\begin{equation}
    H(x; \xi, \sigma, \mu) = 1 - \left( 1 + \frac{\xi(x - \mu)}{\sigma} \right)^{-1/\xi}, \quad x \geq \mu
\end{equation}
where $\mu$ is the threshold, $\sigma>0$ is the scale parameter controlling the spread of excess values, and $\epsilon$ is the shape parameter determining the tail behavior. A positive $\epsilon$ indicates heavy tails with more extreme values, a negative $\epsilon$ shows bounded tails with limited extreme values.
By fitting the tail distribution of observed data, GPD enables simulation of rare but impactful events such as market crashes, extreme rainfall, or infrastructure failures, supporting stress testing and resilience planning in high-risk systems.

\subsection{Generative Models}
\subsubsection{Generative Adversarial Networks (GANs)}

GANs were proposed by Ian Goodfellow \cite{goodfellow2014generative} for generative modeling. It involves two neural networks: the generator $G$ and the discriminator $D$, which engage in a min-max game. The generator aims to produce synthetic data that mimics real data $x$, while the discriminator evaluates whether the data is real or generated. 
\begin{equation}
\min_G \max_D \mathbb{E}_{x \sim p_{\text{data}}(x)}[\log D(x)] + \mathbb{E}_{z \sim p_z(z)}[\log(1 - D(G(z)))]
\end{equation}
Here, $z$ is noise from a prior distribution $p_z(z)$, and $G(z)$ is the generated data. The generator minimizes the discriminator's ability to detect fakes, while the discriminator $D(x)$ maximizes its classification accuracy. 
This adversarial training process leads to the generator learning to create realistic data distributions. 

GANs have seen widespread success in various domains \cite{gu2023stock,gu2025ragic,dash2024hi}. However, they often suffer from training instability, mode collapse, and difficulty in convergence. Many variations and improvements have been proposed to address these issues.
Wasserstein GANs (WGANs) \cite{arjovsky2017wasserstein} replace the original GAN loss with the Wasserstein distance, improving training stability and helping to avoid issues like mode collapse.  It uses a critic instead of a classifier, enforcing the 1-Lipschitz constraint through weight clipping or gradient penalties.
Conditional GANs (CGANs) \cite{mirza2014conditional} extend GANs by conditioning both the generator and discriminator on additional information (e.g., class labels or attributes) for more controlled data generation, such as generating images of specific classes or attributes.
Deep Convolutional GAN (DCGAN) \cite{radford2015unsupervised} applies convolutional neural networks (CNNs) and introduces convolutional architectures to improve the modeling of spatial patterns, producing higher-quality images compared to fully connected networks.

\subsubsection{Variational Autoencoders (VAEs)}
VAEs \cite{kingma2013auto} are probabilistic approaches consisting of an encoder and a decoder. The encoder approximates the posterior distribution $q_\phi(z|x)$, mapping input data $x$ into a latent space, typically assumed to be Gaussian. The decoder reconstructs the input from a latent sample $z \sim q_\phi(z|x)$ and generates data via $p_\theta(x|z)$. VAEs are trained to maximize the evidence lower bound (ELBO) on the data log-likelihood:
\begin{equation}
\mathcal{L}(\theta, \phi; x) = \mathbb{E}_{q_\phi(z|x)}[\log p_\theta(x|z)] - D_{\text{KL}}(q_\phi(z|x) \parallel p(z))
\end{equation}
This objective balances two components: the reconstruction error, encouraging accurate recovery of input from latent space, and the Kullback-Leibler (KL) divergence, regularizing the learned posterior towards the prior (usually a standard normal distribution). 
This framework encourages a smooth and structured latent space, where similar inputs are mapped to nearby latent representations. VAEs are particularly useful for tasks where probabilistic modeling is beneficial. However, they often produce blurry results in high-quality image generation due to the limitations of the Gaussian decoder and the trade-off between reconstruction quality and latent space regularization. Various enhancements have been proposed, including richer priors \cite{tomczak2018vae}, hierarchical latent spaces \cite{sonderby2016ladder}, and hybrid models combining VAEs with GANs \cite{larsen2016autoencoding}.

\subsubsection{Diffusion Models}
Diffusion models \cite{ho2020denoising} have achieved remarkable success in generating high-quality data such as images, audio, and video. They are inspired by the physical diffusion processes, where data points are progressively corrupted with noise over multiple steps. Formally, the forward process gradually perturbs the data $x_0$ over $T$ steps to produce an increasingly noisy version, typically modeled as a Markov chain. The reverse process is learned using a neural network to approximate the posterior $p_\theta(x_{t-1} | x_t)$.
\begin{gather}
q(x_t | x_{t-1}) = \mathcal{N}(x_t; \sqrt{1 - \beta_t} x_{t-1}, \beta_t I)\\
p_\theta(x_{t-1} | x_t) = \mathcal{N}(x_{t-1}; \mu_\theta(x_t, t), \sigma_\theta^2(x_t, t) I)
\end{gather}
Rather than directly learning the full data distribution, the model is trained to predict the noise added at each step, enabling recovery of the original data through iterative denoising. 

While computationally expensive due to the iterative nature of denoising with hundreds or thousands of steps, diffusion models consistently produce highly detailed and realistic outputs. Their training is stable and free from mode collapse, making them a reliable foundation in generative AI. Ongoing research focuses on accelerating inference and improving efficiency to make diffusion models more practical for real-time and large-scale applications.

\section{Methodologies}\label{sec:method}

This section reviews generative modeling approaches specifically designed for extreme event synthesis, grouped into six key categories: (1) EVT-enhanced generative models that modify sampling, training objectives, or latent representations for tail behaviors; (2) importance sampling techniques that bias generation toward extreme regions; (3) diffusion models offering improved stability; (4) hybrid architectures combining multiple paradigms; (5) domain-constrained generative models embedding physical constraints; and (6) LLMs for unstructured or context-rich data generation. Each approach addresses the central challenge of producing realistic synthetic samples that reflect the statistical properties and complexity of rare, high-impact events. Table ~\ref{tab:method} presents a taxonomy of these techniques, mapping each methodological category to its corresponding sections and representative existing work, providing a structured overview of the current landscape in synthetic extreme data generation research.

\begin{table*}

\caption{Taxonomy of techniques for synthetic data generation on extreme data.}\centering\small
\begin{tabular}{m{4cm} m{2cm} m{3cm}}
\toprule
\textbf{Method} & \textbf{Section} & \textbf{Existing Work} \\
\midrule
EVT \& GAN & \ref{sec:411} - \ref{sec:413} & \cite{exgan,gpgan,girard2024deep,huster2021pareto,tailgan,allouche2022ev,boulaguiem2022modeling} \\
EVT \& VAE & \ref{sec:414} & \cite{lafon2023vae} \\
GAN \& different sampling & \ref{sec:411} and \ref{sec:sampling} & \cite{exgan,gpgan,girard2024deep,zhao2024gan}\\
GAN \& different training & \ref{sec:412} - \ref{sec:413} & \cite{huster2021pareto,tailgan,allouche2022ev,boulaguiem2022modeling}\\
Diffusion & \ref{sec:diffusion} & \cite{bassetti2024diffesm} \\
GAN \& VAE & \ref{sec:hybrid} & \cite{azimi2024zgan}  \\
GAN \& physical constraints & \ref{sec:domain_gen} & \cite{karimanzira2024mass,yoon2019time,li2024improved, hongxia2023improved,yi2024method,li2023c}\\
LLMs & \ref{sec:llm} & \cite{disease} \\
\bottomrule
\end{tabular}
\label{tab:method}
\end{table*}

\subsection{EVT-Enhanced Generative AI}

While generative models like GANs and VAEs have shown strong potential for approximating complex distributions, they often focus on high-density regions and tend to produce typical samples near the center of the distribution, resulting in poor representation of extremes. To address this, EVT provides a statistical foundation for modeling rare and heavy-tailed phenomena and has been integrated into generative models to better capture the asymptotic behavior of extremes. These EVT-enhanced models vary in their implementation, including tail-focused sampling, training mechanisms, loss modifications, and latent representations. We organize these methods by their core methodological contribution.

\subsubsection{EVT-Guided Sampling and Tail Conditioning}\label{sec:411}
Instead of generating samples uniformly across the entire data space, these methods use EVT-derived measures, such as extremeness scores or tail quantiles, to bias the sampling process or condition the generator on fitted tail distributions (e.g., GPD \cite{castillo1997fitting}).

ExGAN~\cite{exgan} was among the first to integrate EVT directly into the generative process to explicitly generate samples from tail regions. It introduced an extremeness measure that quantifies how extreme a data point is, and an extremeness probability to describe the quantile level for tail sampling, to fit a GPD. ExGAN drew from learned empirical distributions while conditioning the generator on the fitted GPD to produce samples in the upper tail of the distribution. This framework effectively approximated the EVT-defined asymptotic behavior of extremes and generates realistic yet rare data points, such as extreme precipitation events, without requiring large empirical extreme-event data.

Building on this, Generalized Pareto GAN (GPGAN) \cite{gpgan} extended ExGAN to model multi-dimensional GPD. Unlike prior approaches that assumed identical GPDs across dimensions, GPGAN employed a flexible multivariate GPD to capture extreme occurrences across all dimensions, allowing fine-grained control over the inter-dimensional tail dependencies. Its key innovation was in an adaptive level selection strategy that dynamically determined appropriate GPD parameters and extremeness levels during generation. Specifically, the generator drew from regions defined by a user-specified extremeness function within the GPD framework, then the adaptive strategy adjusted the boundary of the extreme distribution on a per-sample basis. This guarantees more accurate and consistent rare-event generation across samples and dimensions.

HTGAN \cite{girard2024deep} addressed the challenge of modeling heavy-tailed distributions by modifying the latent space of standard GANs. Traditional GANs typically relied on light-tailed latent priors, such as Gaussian distributions, which limit their ability to generate extreme values. HTGAN replaced this latent noise with a heavy-tailed distribution, enabling the generator to more effectively produce extreme values. The model further established theoretical guarantees on its ability to approximate the stable tail dependence function, a core concept in EVT that captures dependence in the tails. Empirical evaluations on both synthetic and real-world heavy-tailed datasets showed that HTGAN significantly outperforms traditional light-tailed GANs in capturing extreme behaviors.

\subsubsection{EVT-Based Training Modifications}\label{sec:412}

Beyond sampling strategies, several methods embed EVT principles directly into the training objective and model architecture to improve the generative model's ability to capture asymptotic tail behavior. These approaches modify loss functions and network parameterizations to reflect heavy-tailed properties, ensuring that the generator aligns with EVT-based theoretical constraints.

Pareto GAN \cite{huster2021pareto} targeted the limitations of standard GANs in modeling heavy-tailed data, which often contained infinite moments and distant outliers. To overcome this, it refined both the generator’s function space and the loss function. Specifically, Pareto GAN integrated EVT to estimate the tail index of the marginal input distributions and incorporated it into the generator to align synthetic data with the tail behavior. It also replaced conventional gradient-based divergence measures with alternative losses defined in tail-aware metric spaces, thus stabilizing training convergence. Pareto GAN was particularly well-suited for extreme-risk financial applications such as credit risk estimation, systemic banking crises, and market stress testing. 

Tail-GAN \cite{tailgan} introduced a GAN framework specifically designed to simulate financial scenarios with accurate tail-risk properties, such as Value-at-Risk (VaR) and Expected Shortfall (ES). Rather than standard adversarial training for tail estimation, Tail-GAN incorporated a custom loss function to exploit the joint elicitability property of VaR and ES, and accurately captured the tail risk of both static and dynamic benchmark portfolios. Furthermore, Tail-GAN established a universal approximation theorem, demonstrating the generator’s ability to approximate a broad class of tail distributions. This made it a promising tool for stress testing and risk management in high-stakes financial applications. 

EV-GAN \cite{allouche2022ev} identified the key issue that feedforward neural networks with ReLU activations poorly approximated unbounded quantiles of heavy-tailed distributions. 
It solved this issue by a new parametrization of the GAN that transformed the problematic quantile function of heavy-tailed distributions into a bounded tail-index function that neural networks can effectively approximate. The generator learned this transformed function and applied the inverse transformation to produce properly distributed extreme events. Their theoretical analysis showed the promising approximation error's convergence rate. This approach was particularly necessary in financial domains where outliers, such as market crashes or liquidity crises, must be modeled with high reliability for effective risk management and stress testing.

\subsubsection{Structured and Spatial Tail Modeling}\label{sec:413}

Rare events often exhibit dependencies across space and time, which standard EVT models may fail to capture.
EVT-GAN \cite{boulaguiem2022modeling} addressed the challenge of modeling spatial extremes. It combined DCGANs with multivariate EVT, especially to approximate the spatial dependence among rare event observations. Exploiting EVT's asymptotic results for daily maxima, the model learned a joint latent representation that captures both spatial co-occurrence and tail correlations across variables such as wind speed, significant wave height, and total precipitation. This enabled more realistic simulation of rare events with geographically correlated impacts.

\subsubsection{Latent Decomposition}\label{sec:414}

While GANs dominate EVT-integrated research, VAEs offer advantages in training stability, likelihood-based learning, and interpretability. 
ExtVAE \cite{lafon2023vae} was the first to integrate EVT into the VAE framework to synthesize multivariate extremes in heavy-tailed distributions. 
Conventional VAEs typically assume Gaussian priors and thus struggle to capture tail dependencies and perform well under data sparsity. ExtVAE addressed this by leveraging EVT to provide the theoretical foundation through multivariate regular variation, which decomposed each data sample into a radius (representing event magnitude or extremeness) and an angular measure (capturing the direction or dependence structure among variables when extremes occur). The radius was modeled using a univariate VAE with a heavy-tailed inverse-Gamma prior and a decoder constrained to maintain heavy-tailed behavior via tailored neural parameterizations. The angular component was generated via a conditional multivariate VAE on the unit simplex conditioned on the radius, with an enforced asymptotic independence between angle and radius, as required by EVT. This design enabled ExtVAE to accurately represent both marginal tail behavior and complex joint extremes.

\subsection{Importance Sampling Integration}\label{sec:sampling}

One core challenge in rare-event modeling is the scarcity of extreme examples in training data, making it difficult for generative models to learn tail distributions effectively. Importance sampling offers a practical solution by biasing the latent space toward extreme regions, guiding the generator to focus on underrepresented but high-impact samples.
Gaussian Mixture Model-GAN (GMM-GAN) \cite{zhao2024gan} combined a GMM-based importance sampling with GAN architectures. Instead of using a standard latent prior, it sampled latent vectors from a GMM, which is biased toward extreme regions of the input distribution. Each generated scenario was paired with a corresponding sampling weight that reflected its importance. This sampling bias guided the training process of GAN to prioritize tail-event generation without disrupting the overall distribution. GMM-GAN not only replicated the empirical distribution of renewable power data accurately but also enhanced the efficiency and diversity of synthetic tail-event samples.
Importance sampling thus offered a practical solution for rare-event targeting without altering the underlying model architecture or requiring large volumes of extreme data.

\subsection{Emerging Diffusion Models}\label{sec:diffusion}

Diffusion models provide a fundamentally different mechanism for data generation. Instead of mapping noise to data directly, they learn to reverse a noise diffusion process through a score function—the gradient of the data log-density. They are known for their training stability, high sample diversity, and ability to model high-resolution data without mode collapse, making them promising candidates for rare-event synthesis.
DiffESM \cite{bassetti2024diffesm} was a 3D diffusion-based framework inspired by stochastic differential equations, designed to emulate Earth System Model (ESM) outputs with high temporal fidelity. Classical ESMs typically operated at coarse resolutions (e.g., monthly), which limited their utility for modeling short-duration extreme weather events. 
DiffESM addressed this by training a diffusion model on paired monthly and daily climate data to learn the rescaling process from coarse to high-resolution sequences. This conditional generation process accurately reconstructed daily extremes, such as precipitation and temperature, while preserving spatial correlation structures inherent in the ESM outputs.
By enabling high-resolution reconstruction of climate extremes, DiffESM demonstrated the potential of diffusion models for downscaling and rare-event modeling for climate risk assessment. These advances highlighted diffusion models as a stable and high-fidelity alternative for generating extreme sequences in environmental systems.

\subsection{Hybrid Generative Models}\label{sec:hybrid}
Hybrid architectures combine multiple generative paradigms to leverage their strengths in modeling complex or rare distributions. In particular, integrating VAEs and GANs can balance the stability and explainability of VAEs with the realism of GAN-generated samples.
zGAN \cite{azimi2024zgan} exemplified this approach by targeting outlier generation for tabular data, with applications in financial and risk modeling. It integrated a conditional VAE \cite{sohn2015learning} and a GAN within a unified architecture. The conditional VAE learned structured latent representations of real data, generated synthetic covariance matrices to preserve statistical properties, and produced outliers that exhibited rare or extreme properties. The generator in GAN produced synthetic data from random noise and applied a hash-based similarity filter to protect privacy. By augmenting generated data with outliers, zGAN created high-fidelity, diverse, and explainable synthetic tabular datasets with controlled extreme event representations, while preserving data utility and privacy.

\subsection{Domain-Constrained Generative AI}\label{sec:domain_gen}

In scientific and engineering domains, rare events are often governed by underlying physical, economic, or operational constraints. Standard generative models may produce unrealistic samples when these domain-specific laws are not embedded. To overcome this, several approaches explicitly incorporated domain knowledge into the generative process through conditioning mechanisms, auxiliary inputs, or regularization.

Mass-Conservative Time-Series GAN (MC-TSGAN) \cite{karimanzira2024mass} modified TimeGAN \cite{yoon2019time}, for flood-event generation by introducing hydrology principles. It embeds conservation laws, such as mass and energy balance, as regularization terms directly into the loss function. The model leveraged LSTM-based generators and discriminators to capture temporal dependencies, and applied a space-filling sampling strategy to reduce large biases and improve data quality. Dimensionality reduction techniques such as PCA and t-SNE confirmed improved structure fidelity in synthetic data, and \textit{t}-tests validated statistical similarity to real data. When integrated with real-world observation data, synthetic flood-event data significantly improved the robustness and accuracy of the probabilistic neural runoff model for multi-step flood forecasting, demonstrating the practical utility of physics-aware generative modeling in hydrological extremes.

In the energy domain, where rare-event modeling targets scenarios such as prolonged storms or surges, conditional GANs have been widely used \cite{li2024improved, hongxia2023improved}. 
InfoGAN \cite{yi2024method} applied mutual-information-based latent structure learning to extract and generate solar-wind-load scenarios with preserved correlation structures under extreme demand conditions, accounting for spatiotemporal dependencies that standard GANs overlook. 
Similarly, \cite{li2024improved} employed conditional GANs for extreme renewable energy scenarios, utilizing Wasserstein loss to generate photovoltaic power outputs under extreme weather conditions.
Conditional DCGAN (C-DCGAN) \cite{li2023c} integrated conditional inputs from weather indicators into a DCGAN framework, using Wasserstein loss and a specialized ranking mechanism to select the most extreme energy risk scenarios from generated samples. It enabled the generation of rare yet plausible scenarios across the joint distribution of renewable power generation, storage, and demand, such as long-duration storms or grid failures.

By embedding domain-specific constraints and knowledge, these models ensured that synthetic extreme events remain physically meaningful, statistically plausible, and applicable to high-impact decision-making.

\subsection{Large Language Models (LLMs)}\label{sec:llm}

Large Language Models (LLMs) offer a novel and flexible approach to rare-event simulation, particularly in domains where data is sparse, sensitive, or unstructured. Unlike traditional generative models focused on numerical or image data, LLMs are trained on massive textual corpora and can generate rich, semantically grounded outputs, making them well-suited for simulating extreme scenarios involving human behavior, decisions, and system interactions.
Recent studies have explored the use of LLMs in the healthcare domain, where data scarcity and privacy concerns limit access to real patient records. One study~\cite{disease}, though not proposing a specific model, investigated how LLMs and other generative AI techniques could synthesize clinical narratives for rare disease research. These models can preserve essential statistical and linguistic properties of medical records while ensuring patient privacy through differential privacy and federated learning. This capability opens new avenues for collaboration and model development in data-constrained healthcare settings.

% \textbf{Narrative-Driven Simulation from Fiction and Reflection}:
% Building on this foundation, we propose a promising research direction that leverages narrative-based corpora, which are often excluded from structured datasets, to synthesize extreme scenarios with greater nuance and realism. 
% Fictional disaster narratives, such as those found in speculative fiction, novels, films, and games, often describe complex, multi-system failures, emergent threats, and societal responses that have not yet occurred in reality. By learning from these sources, LLMs can simulate edge-case scenarios that challenge existing risk assumptions and broaden the range of possibilities considered in planning and forecasting.
% Complementing this, reflective accounts from real responders, such as after-action reports, memoirs, oral histories, and expert interviews, provide critical insight into expert decision-making, adaptive strategies, and moments of uncertainty during actual crises. Incorporating such reflections would enable LLMs to generate human-centered simulations that capture the lived experiences of rare events.

\section{Data}\label{sec:data}

This section provides an overview of benchmark datasets commonly used in extreme synthetic data generation research. We categorize these datasets into five major types: image data, time series data, graph data, tabular data, and other forms of data, each specifically tailored to different extreme event scenarios and modeling challenges.
For each category, we introduce the characteristics of each data type, followed by representative datasets with their sources, resolution, and extreme event definitions. We emphasize how each dataset captures rare phenomena and highlight their applications in existing research.
Table \ref{tab:data} and Table \ref{tab:data_small} provide structured summaries of these datasets, presenting key metadata including data sources, extreme case focus, temporal coverage, and  availability\footnote{https://github.com/JingyiGu/synthetic-rare-event-benchmark}. These tables serve as practical references for researchers seeking appropriate benchmarks for their specific extreme event synthesis tasks, facilitating reproducible research.

\begin{table*}
\centering
\caption{Benchmark datasets for synthetic data generation for rare events. A star {*} indicates the dataset is publicly available.}
\resizebox{\textwidth}{!}{%
\begin{tabular}{m{1.7cm}m{4cm}m{7cm}m{6cm}}
\toprule
\textbf{Type} & \textbf{Dataset} & \textbf{Description} & \textbf{Extreme-event focus} \\
\midrule
Image & US Precipitation *~\cite{exgan,gpgan} & Hourly 2-D precipitation heatmaps on a 4 km × 4 km grid & Extreme rainfall for top daily 95$^{\text{th}}$ percentile. \\
\midrule
\multirow{7}{1.7cm}{Finance Time Series} 
 & Nasdaq ITCH *~\cite{tailgan} & Limit order book data for high-frequency trading, microsecond-stamped level-3 orders for Nasdaq stocks & Heavy-tailed bid/ask prices with market volatility for flash-crash studies \\
\cmidrule(l){2-4}
 & Global Indices *~\cite{allouche2022ev} & Daily index prices from global markets & Tail dependence in log returns \\
\cmidrule(l){2-4}
 & Kaggle S\&P 500 *~\cite{girard2024deep} & Daily OHLCV for 500 constituents in 2010-2024 & Sector-specific co-movement tails \\
\cmidrule(l){2-4}
 & QuantQuote S\&P 500 *~\cite{huster2021pareto} & Long-history intraday and daily OHLCV for S\&P 500 stocks & Stock-level heavy tails \\
\cmidrule(l){2-4}
 & Simulated Return *~\cite{tailgan} & Synthetic Gaussian, AR, and GARCH process for return series & Designed heavy tails \& complex volatility \\
\cmidrule(l){2-4}
 & Copula Returns~\cite{allouche2022ev} & Synthetic bivariate/multivariate returns from Gumbel copulas & Upper-tail dependence \\
\cmidrule(l){2-4}
 & Market Supervision *~\cite{jiang2024leveraging} & SEC filings, TRACE Bond transactions, FRED macroeconomic indicators & Fraud, market crash, or system risk detection \\
\midrule
\multirow{4}{1.7cm}{Climate Time Series}
 & CESM *~\cite{bassetti2024diffesm} & Global daily climate simulations for atmospheric, oceanic, land, sea ice component systems in 1920–2100 & Internal variability (temperature, precipitation, etc), \& extremes in future climate \\
\cmidrule(l){2-4}
 & IPSL *~\cite{bassetti2024diffesm} & Monthly/daily global climate simulation in 1850–2300 & Extreme weather events for multi-model comparison. \\
\cmidrule(l){2-4}
 & EC-Earth v2.3 *~\cite{boulaguiem2022modeling} & 2000-year daily hydro-climate simulation & Regime shifts \& long-term extreme event attribution \\
\cmidrule(l){2-4}
 & ERA5 *~\cite{peard2023combining} & Hourly climate observation for global atmosphere, land, surface, and waves in 1940–2022. & Hazard simulation \\
\midrule
\multirow{4}{1.7cm}{Renewable Energy Time Series}
 & Southern Power Grid~\cite{hongxia2023improved} & 15-min Wind \& PV across 5 provinces & Energy shortfall events \\
\cmidrule(l){2-4}
 & Northwestern Energy~\cite{li2023c} & 5-min Wind \& PV in one year & Severe outage or low-generation \\
\cmidrule(l){2-4}
 & Belgian Grid *~\cite{yi2024method} & 15-min wind, solar, load \& weather data in 2018–2022. & Renewable-load imbalance for grid operation and planning \\
\cmidrule(l){2-4}
 & GEFCom2014 *~\cite{zhao2024gan} & Hourly electricity load, price, wind, solar in 731 days & High wind power days $\ge$ 60–90\% capacity \\
\midrule
\multirow{2}{1.7cm}{Hydrology}
 & Germany Ahrtal Floods~\cite{karimanzira2024mass} & 10–15 min water levels, flow rates, precipitation, soil information from multiple gauges & Severe flood simulation \\
\cmidrule(l){2-4}
 & Danube River *~\cite{lafon2023vae} & Daily river discharge from 31 stations in 50 years & High-flow tails with temporal and spatial dependence \\
\midrule
\multirow{2}{1.7cm}{Event / Behavior Time Series}
 & Keystrokes *~\cite{huster2021pareto} & 136m inter-keystroke events for 168k users & Unusual typing dynamics, e.g., long/short delays \\
\cmidrule(l){2-4}
 & Wikipedia Traffic *~\cite{huster2021pareto} & Daily page-views for 145k Wikipedia articles over 550 days & Web traffic spike extremes \\
\midrule
Graph & SNAP LiveJournal *~\cite{huster2021pareto} & Social networks with 4m nodes/users and 34m edge friendship & Heavy-tailed node degree with hub extremes \\
\midrule
\multirow{2}{1.7cm}{Tabular}
 & 2-D Gaussian Mixture Toy~\cite{zhao2024gan} & Two-dimensional Gaussian mixture distribution with 8 centers & Low-probability tails on one dimension \\
\cmidrule(l){2-4}
 & 5-D Heavy-tail Toy~\cite{lafon2023vae} & 5-D Dirichlet-angular heavy-tail distributions with sparse training set & Rare-event generation on limited samples \\
\bottomrule
\end{tabular}}
\label{tab:data}
\end{table*}

\begin{table*}[htbp]

\caption{Real-world vs. simulated datasets.}\centering\small
\begin{tabular}{m{3.3cm} m{8cm}}
\toprule
 \textbf{Source Type} & \textbf{Datasets} \\
\midrule
Real-World Data & \cite{allouche2022ev, boulaguiem2022modeling, bassetti2024diffesm, exgan, girard2024deep, gpgan, hongxia2023improved, huster2021pareto, jiang2024leveraging, karimanzira2024mass, lafon2023vae, li2023c, peard2023combining, tailgan, yi2024method, zhao2024gan}
 \\\midrule
Simulated Data & \cite{allouche2022ev,lafon2023vae,zhao2024gan,tailgan} \\
\bottomrule
\end{tabular}
\label{tab:data_small}
\end{table*}

\subsection{Image Data}
The \textbf{US precipitation dataset} provides daily precipitation measurements on a spatial grid covering the Continental United States, Puerto Rico, and Alaska. 
Each data sample is presented as a 2D heatmap generated from a multi-sensor approach, representing hourly observed precipitation at a spatial resolution of approximately $4\times4$ km. The dataset spans from January 2010 to August 2020 and is publicly accessible via the National Weather Service\footnote{https://water.weather.gov/precip/}.
It has been used extensively for advanced extreme weather modeling, such as ExGAN 
\cite{exgan} and Generalized Pareto GAN \cite{gpgan}, which utilized different temporal segments and preprocessing techniques to generate synthetic precipitation data that preserve the statistical characteristics of extreme weather events. 
To effectively highlight extreme precipitation events, the test set exclusively includes samples with total rainfall exceeding the 95th percentile of the training data. Original heatmaps at a resolution of $813 \times 1051$ pixels are resized to $64\times64$ pixels and normalized within the range $[-1,1]$.
This dataset supports spatially structured rare-event generation and has become a standard benchmark in climate-related synthetic data research.

\subsection{Time Series Data}
\subsubsection{Financial Datasets}
The financial sector provides rich time series data widely used for extreme data generation research. These datasets capture detailed temporal patterns essential for studying market dynamics, volatility, and systemic financial risks.

\paragraph{a. Limit Order Book}
The limit order book (LOB) provides a detailed view of buy and sell orders across various price levels. It reflects real-time market supply-demand dynamics and plays a key role in determining market prices. Bid and ask prices derived from the best available buy and sell orders serve as important indicators of market conditions. During extreme financial events such as flash crashes or liquidity shocks, these prices exhibit significant deviations. Focusing on the tail of the return distribution of order price data enables the analysis of market volatility, liquidity crises, and other high-impact events beyond typical market movements.
\textbf{Nasdaq ITCH data}, available through LOBSTER\footnote{https://lobsterdata.com/}, 
is a high-frequency benchmark containing level-3 LOB data with microsecond precision for Nasdaq-listed stocks. This dataset captures every order-related event, including submissions, modifications, and cancellations, along with key features such as order book depth (with top bid and ask levels), prices, direction, and size. It provides a granular view of market dynamics, including periods of high volatility and extreme price movements, ideal for modeling extreme market conditions, as demonstrated by Tail-GAN \cite{tailgan}.
Specifically, the mid-price series (the average of bid and ask prices) of five prominent stocks is sampled at a 9-second interval. Each sampled series consists of 100 data points over a 15-minute interval, allowing Tail-GAN to capture the heavy-tailed distribution of mid prices and synthesize market extreme scenarios.

\paragraph{b. Global Market Indices}
Global market indices provide aggregated representations of market behaviors across multiple sectors, economies, and regions, essential for capturing systemic and correlated financial risks during global crises \cite{gu2023deep}. 
Their high liquidity and diversified structure make them ideal for studying large-scale extreme market swings, while reducing susceptibility to localized noise or market manipulation. 
The global indices are typically publicly available through platforms such as Yahoo Finance\footnote{https://finance.yahoo.com/markets/world-indices/},
which lists major indices from the Americas, Europe, and Asia-Pacific regions, and Stooq\footnote{https://stooq.com/db/h/},
which includes historical data on 61 world indices dating back to their inception. In EV-GAN \cite{allouche2022ev}, six geographically diverse indices, NKX (Japan), KOSPI (Korea), HSI (Hong Kong), CAC (France), AMX (Netherlands), and Nasdaq (USA), are employed to compute daily log-returns to train generative models for time series data synthesis that accurately reflect tail and dependence properties.

\paragraph{c. Price Data for Individual Assets}
Individual asset prices are crucial for modeling extreme market scenarios, as they directly reflect idiosyncratic risks and sector-specific shocks. Asset-level data captures fine-grained fluctuations and localized anomalies, enabling generative models to learn nuanced patterns for extreme value analysis.

\emph{c.1. Real Price Data}.
Real-world collected price data provides a crucial foundation for capturing authentic market dynamics and extreme behaviors in financial modeling. The S\&P500 is a key benchmark, tracking the 500 most valuable publicly traded U.S. companies. Two widely used datasets provide high-quality price series for individual S\&P500 stocks. (1) The \textbf{Kaggle S\&P 500 dataset}\footnote{https://www.kaggle.com/datasets/camnugent/sandp500} 
offers daily OHLCV (Open, High, Low, Close, Volume) data for all companies listed in the index from 2010 to 2024 across multiple sectors. It is utilized in HTGAN \cite{girard2024deep} to examine sector-specific tail behaviors, particularly in the Financial (27 tickers) and Utilities (57 tickers) sectors, both prone to abrupt regime shifts and volatility spikes during economic stress. 
(2) \textbf{QuantQuote historical data}\footnote{ https://quantquote.com/historical-stock-data} 
delivers intraday resolution prices with time intervals ranging from 1 minute to 1 day, along with full daily OHLCV records dating back to 1999. This dataset is used in Pareto-GAN \cite{huster2021pareto} for developing robust models capable of generating heavy-tailed behaviors in real markets.

\emph{c.2. Simulated Price Data}.
In addition to real data, simulated return datasets provide controlled environments for systematically evaluating generative performance. These synthetic benchmarks are designed with explicit statistical properties, allowing models to be tested for their ability to replicate autocorrelation, volatility clustering, and heavy-tailed distributions. (1) One benchmark consists of \textbf{five simulated financial assets}\footnote{https://github.com/chaozhang-ox/Tail-GAN}, each exhibiting different temporal dynamics and tail behaviors. The marginal distributions contain a Gaussian distribution, two AR(1) processes with opposite autocorrelation patterns, and two GARCH(1,1) processes incorporating \textit{t}-distributed noise with varying degrees of freedom. This diversity enables fine-grained evaluation of a model's capability to reproduce tail distributions and complex volatility dynamics in Tail-GAN \cite{tailgan}.
(2) Another benchmark is a \textbf{Copula-based return data} \cite{allouche2022ev} in both bivariate and multivariate settings. These are constructed using parametric copula models, separating marginal distributions from the dependence structure. The dependence structures are modeled using Gumbel copulas \cite{wang2010using} with varying dependence parameters to represent upper-tail dependence, while the margins follow Burr distributions \cite{fry1989univariate} with varying tail indices and second-order parameters. 
The simulated series is used in EV-GAN\cite{allouche2022ev} to evaluate tail dependencies and joint extreme behaviors in higher-dimensional synthesis.

\paragraph{d. Financial Market Supervision}
Rare-event modeling is particularly critical in the context of financial market supervision, where the detection of fraud, systemic risk, and market manipulation depends on subtle and infrequent signals. A comprehensive, multi-source dataset has been assembled to support research in this domain, integrating records monitored by key U.S. regulatory agencies.
This dataset incorporates public company filings and financial statements from the SEC EDGAR Database\footnote{https://www.sec.gov/search-filings/edgar-search-assistance/accessing-edgar-data}
, detailed bond transaction records from FINRA TRACE\footnote{https://www.finra.org/filing-reporting/trace/data}
, and macroeconomic indicators from the Federal Reserve Economic Data (FRED) platform\footnote{https://fred.stlouisfed.org/}. As detailed in \cite{jiang2024leveraging}, this dataset consists of thousands of samples, each representing a unique financial entity or transaction, and includes over 20 features such as transaction volume, market volatility, financial health, and regulatory disclosures. This integration of heterogeneous data sources enables the dataset to reflect both firm-level and market-wide financial behaviors. 
A notable characteristic is its class imbalance: high-risk events (e.g., fraud or market crashes) occur far less frequently than normal market activity. This imbalance poses significant challenges for traditional supervised learning models to accurately identify rare but critical signals. Therefore, this data is a reliable resource for synthesizing high-risk samples to augment training data and improve model robustness. Its breadth and complexity make it valuable for advanced risk prediction systems in financial market regulation \cite{jiang2024leveraging}.

\subsubsection{Climate Datasets}
Climate data is essential for modeling rare and extreme environmental events, offering long-term, multivariate, and globally distributed information. These datasets typically are spatiotemporal time series containing variables such as temperature, precipitation, wind, and sea-level pressure in long durations. 

ESM simulations provide physically consistent climate outputs under historical and future scenarios. 
The Community Earth System Model \textbf{CESM} dataset\footnote{https://www.cesm.ucar.edu/community-projects/lens/data-sets}, developed by the NCAR-DOE Community ESM \cite{kay2015community}, provides long-term simulations to capture how internal variability shapes long-term climate trajectories. It includes 30 ensemble simulations from 1920 to 2100 with slightly perturbed initial conditions. It simulates interactions across the atmospheric, oceanic, land, and sea ice systems. Daily variables such as temperature, precipitation, and sea ice concentration, across both time and space, support research on internal variability, extremes, and future climate projections.

The Institut Pierre-Simon Laplace (\textbf{IPSL}) dataset\footnote{https://data.ipsl.fr/catalog/srv/eng/catalog.search\#/home}, derived from the IPSL model \cite{dufresne2013climate}, extends this timeline from the pre-industrial period (1850) to future scenarios (2300). It provides monthly and daily variables like temperature, precipitation, pressure, and wind, on a coarser spatial resolution. Though structurally similar to CESM, it reflects different modeling assumptions and physical parameterizations, making it valuable for multi-model comparisons and climate feedback analysis on extreme weather events. These two datasets are used in DiffESM \cite{bassetti2024diffesm}.

\textbf{EC-Earth (v2.3)} dataset \cite{van2019added}, derived from EC-Earth global climate model \cite{hazeleger2012ec}, simulates 2000 years of daily climate data across atmospheric, ocean, land surface, and sea ice systems. It is designed to explore the impact of natural variability on extreme hydrological events under both historical and scenario-based conditions. The long temporal span and detailed climate dynamics make it ideal for studying regime shifts, variability, and extreme event attribution \cite{boulaguiem2022modeling}. 

The \textbf{ERA5}\footnote{https://cds.climate.copernicus.eu/cdsapp\#!/dataset/reanalysis-era5-single-levels} 
\cite{hersbach2020era5} reanalysis data, provides hourly records from 1940 to 2022 by blending observational data with numerical weather prediction models. It includes the global atmosphere, land surface, and ocean waves, capturing fine-scale weather pattern evolution. ERA5 is widely used for climate monitoring and validation, and has been applied in spatially consistent synthetic hazard simulations \cite{peard2023combining}.

In summary, we collect and organize representative climate datasets that capture different aspects of extreme event modeling, from long-term variability to short-term weather anomalies, and from physically grounded simulations to data-assimilated reanalysis. By assembling these informative and complementary datasets, we provide a consolidated resource that facilitates future research on synthetic climate data generation.

\subsubsection{Renewable Energy Datasets}

Extreme events in renewable energy systems, such as sudden power ramps and generation shortfalls, pose critical challenges for forecasting, grid reliability, and planning. Several datasets have been used in recent studies to support the synthesis of rare but impactful scenarios in wind and solar energy. 

The \textbf{China's Southern Power Grid Data} provides 15-minute resolution measurements of wind and photovoltaic (PV) power generation data in five provinces of southern China, capturing joint temporal dynamics of wind and solar energy output under real operational conditions. In C-MCDCGAN-GP\cite{hongxia2023improved}, test samples are filtered to retain only the most severe joint low wind–PV cases, simulating high-risk energy shortfalls. 

The \textbf{Northwestern China Renewable Energy Data} offers even finer granularity, with 5-minute resolution wind power and PV generation data collected over 366 days. Each day consists of 288 time steps, yielding over 100,000 data points in total. Designed for modeling extreme low-output scenarios with electricity supply tension, each daily sample is labeled based on the severity of wind and solar underproduction. This enables models to conditionally synthesize realistic high-risk scenarios, supporting for stress testing stress testing of grid reliability under rare but operationally critical power shortages \cite{li2023c}. 

The \textbf{Belgian grid dataset}, adopted in InfoGAN\cite{yi2024method} study, includes historical wind and solar power generation, electricity load, and corresponding meteorological variables such as temperature, wind speed, and solar irradiance. Energy data is sourced from the Belgian grid\footnote{https://www.elia.be/en/grid-data}, while weather data is obtained from Wunderground\footnote{https://www.wunderground.com}. It spans from 2018 to 2022 at 15-minute intervals and contains 96 records per day per variable. Its multi-source, high-resolution structure enables modeling of complex dependencies between weather conditions, renewable generation, and load demand, supporting synthetic scenario generation of renewable-load imbalances under uncertainty.

\textbf{GEFCom2014 Data} is a benchmark \cite{hong2016probabilistic} in energy forecasting research. It comprises 731 days of hourly time series covering electric load, electricity price, wind power, and solar. In GMM-GAN\cite{zhao2024gan}, it was leveraged to generate synthetic high-output wind power scenarios. Extremes were defined based on whether daily average wind output exceeded specific percentage thresholds (e.g., 60\%, 80\%, 90\%), enabling the model to conditionally synthesize high-output wind power days for robust scenario planning.

In summary, we curate and present a diverse set of renewable energy datasets that span different temporal resolutions, geographic regions, and renewable sources. This organized collection provides a practical foundation for developing and evaluating synthetic models targeting rare but critical energy scenarios.

\subsubsection{Hydrological Datasets}

Hydrological data is particularly relevant for modeling rare and high-impact flood events due to the heavy-tailed distribution of river flow values \cite{katz2002statistics}. \textbf{Germany's Ahrtal Region Flood Data} describes severe flooding events in the Ahrtal Region. 
It integrates multi-source hydrological data from three strategically located gauge stations, including water levels, river flow rates, and precipitation, detailed soil characteristics (e.g., texture, permeability), water-holding capacity for runoff generation, and water-level measurements in rivers for real-time hydrological conditions, The data is collected at a 10–15 minute intervals, ensuring fine-grained simulation that reflect the complexity and variability of extreme flood events. This comprehensive dataset supports both flood forecasting and disaster risk mitigation \cite {karimanzira2024mass}.
\textbf{The Danube river dataset}\footnote{http://www.gkd.bayern.de} 
consists of 50 years of daily river discharge measurements from 31 hydrological stations in the upper Danube basin. Its multivariate structure captures spatial dependencies across the river network, enabling generative models like VAEs \cite{lafon2023vae} to simulate joint extremes in discharge across locations. The extensive temporal coverage and spatial resolution make the Danube dataset a benchmark for generating synthetic hydrological extremes with real-world complexity.

\subsubsection{Other Time Series Data}
The \textbf{Keystroke Dataset}\footnote{https://userinterfaces.aalto.fi/136Mkeystrokes/}
\cite{dhakal2018observations} comprises over 136 million keystroke events from 168,000 users. Each event is represented by the inter-arrival time between consecutive keystrokes, capturing fine-grained user behavior in typing dynamics. The dataset is primarily used to model tail events, such as unusually long or short inter-keystroke delays, which are rare but significant for applications like anomaly detection, behavioral biometrics, or human-computer interaction analysis. In Pareto-GAN \cite{huster2021pareto}, these inter-arrival times serve as the basis for learning and generating synthetic samples from the extreme regions of the distribution, demonstrating the model’s capacity to replicate atypical human typing behavior.
\textbf{Wikipedia Web Traffic data}\footnote{https://www.kaggle.com/c/web-traffic-time-series-forecasting} \cite{huster2021pareto} 
comprises daily page view counts for approximately 145,000 Wikipedia articles over 550 days. Its large scale and temporal diversity allow for modeling extreme user attention patterns, including traffic surges and long-tail access distributions, making it useful for testing synthetic models on real-world web behavior extremes.
% Pareto-GAN for extreme value generation research.

\subsection{Graph Data}
One category of graph-structured data often exhibits a power-law degree distribution \cite{rodriguez2022arachne}, where a few nodes (hubs) have exceptionally high degrees while the vast majority have only a few connections. This heavy-tailed pattern is prevalent across real-world systems, such as social, biological, and communication networks, and modeling these structural extremes is essential for understanding network robustness, influence dynamics, and rare-event behavior. Capturing and synthesizing such heavy-tailed distributions remains a key challenge in graph generation.
The \textbf{SNAP LiveJournal} dataset \footnote{https://snap.stanford.edu/data/soc-LiveJournal1.html}\cite{leskovec2009community,leskoves2018} is a large-scale social network graph derived from the LiveJournal blogging platform. It contains over 4 million nodes (users) and 34 million directed edges, where each edge indicates a social connection or friendship, forming a complex and sparsely connected network. The dataset is notable for its heavy-tailed node degree distribution, which makes it a canonical example for studying extreme connectivity patterns in complex networks. In Pareto-GAN \cite{huster2021pareto}, node degree values are used as the target variable for modeling and generating synthetic degree distributions with extreme values. The resulting synthetic data reflects the rare but influential high-degree nodes, supporting research in graph-based anomaly detection, influence modeling, and network risk analysis.

\subsection{Tabular Data}

Tabular data consists of fixed-length feature vectors, but lacks the inherent temporal or spatial structure in time series or spatial data. To our knowledge, all tabular datasets in extreme data modeling are synthetically generated rather than derived from real-world sources. They are often used to validate generative models under controlled statistical conditions. Their flexibility allows researchers to explicitly define marginal distributions, dependence structures, and tail behaviors.
One example is \textbf{2D Gaussian mixture toy dataset} \cite{zhao2024gan}. This synthetic dataset consists of a two-dimensional Gaussian mixture distribution with eight centers, with rare events defined through a tail condition on one variable. 
Though simple, it effectively proves how conditional generative models explore low-probability, multimodal regions in the latent space before tackling more challenging real-world data with extreme characteristics.
Another illustrative case is \textbf{5-dimensional synthetic dataset}. It involves a 5-dimensional heavy-tailed random variable with a radius distribution with tail index parameters and an angular distribution following a Dirichlet distribution with radius-dependent parameters, creating complex dependencies between variables. The dataset is intentionally sparse to simulate real-world challenges in extreme event analysis: 250 training samples, 750 for validation, and a comprehensive test set of 10,000 samples. This configuration specifically tests models' ability to accurately extrapolate beyond observed data and generate realistic extreme events with limited training examples, addressing a critical gap in traditional statistical approaches when handling heavy-tailed phenomena. This benchmark dataset was utilized in evaluating the VAE-based approach\cite{lafon2023vae}.
Both datasets serve as domain-agnostic testbeds to evaluate the theoretical robustness of generative models in rare-event synthesis under complex statistical constraints.

\begin{table*}

\caption{Evaluation category and corresponding metrics for synthetic extreme data generation.}\centering\small
\begin{tabular}{m{5.3cm} m{5cm} m{2.2cm}}
\toprule
\textbf{Evaluation Category} & \textbf{Metrics} & \textbf{Existing Work} \\
\midrule
Distributional Similarity (\ref{sec:6.1}) & KS Test, t-statistic, KL Divergence, Wasserstein Distance, FID & \cite{huster2021pareto,gpgan,zhao2024gan,karimanzira2024mass,lafon2023vae} \\\midrule
Dependence Preservation (\ref{sec:6.2}) & AKE, Pearson Correlation, ACF & \cite{girard2024deep, allouche2022ev, yi2024method,hongxia2023improved, tailgan} \\\midrule
Extreme Coverage (\ref{sec:6.3}) & Coverage Rate, Interval Width & \cite{yi2024method, zhao2024gan} \\\midrule
Extreme Dependence (\ref{sec:6.4}) & Extremal Coefficient, Extremal Correlation Coefficient & \cite{peard2023combining,boulaguiem2022modeling} \\\midrule
Reconstruction Loss (\ref{sec:6.5}) & Instance-Wise Reconstruction Loss & \cite{exgan} \\\midrule
Extreme Magnitude Accuracy (\ref{sec:6.6}) & MAPE, RMSE, MSLE & \cite{yi2024method,tailgan,allouche2022ev} \\\midrule
Visualization Diagnostics (\ref{sec:6.7}) & PCA, t-SNE, Q-Q Plot, Rank-Frequency Distribution & \cite{karimanzira2024mass,peard2023combining, lafon2023vae,tailgan} \\\midrule
Downstream Performance Validation (\ref{sec:6.8})& Task-specific metrics & \cite{jiang2024leveraging, karimanzira2024mass} \\
\bottomrule
\end{tabular}
\label{tab:evaluation_metrics}
\end{table*}

% \vspace*{-\baselineskip} 
\section{Evaluation Metrics}\label{sec:metric}

Unlike conventional synthetic data evaluation that primarily emphasizes privacy preservation and general distributional similarity, evaluating synthetic extreme data requires fundamentally different approaches. Our contributions in developing a comprehensive evaluation framework include:
\begin{enumerate}
    \item \textbf{Extreme-focused evaluation framework}: We develop metrics specifically tailored for extreme event synthesis, focusing on capturing rare, high-impact behaviors rather than privacy preservation .

    \item \textbf{Comprehensive metric categories}: Our framework encompasses distributional similarity, dependence preservation, extreme coverage, extremal dependence, reconstruction/prediction quality, and downstream performance validation, along with visualization techniques.

    \item \textbf{Detailed theoretical and practical guidance}: We provide not only conceptual introductions but also detailed mathematical definitions and explicit guidance on applying these metrics to extreme event applications.
    
    \item \textbf{In-depth insights on extremeness}: As our most distinctive contribution, for each metric, we analyze (1) its application to extreme data assessment with advantages, limitations, and trade-offs, and (2) its domain-specific adaptations. This structured comparison goes beyond standard evaluation, providing practical guidance for selecting appropriate metrics tailored to extreme event modeling needs.
\end{enumerate}

Table \ref{tab:evaluation_metrics} presents taxonomy of these evaluation approaches, mapping each category to its specific metrics and applications.

\subsection{Distributional Similarity}\label{sec:6.1}
To evaluate the fidelity of synthetic data, particularly in replicating the statistical structure of extreme events, it is essential to match marginal distributions between real and synthetic data. Various statistical metrics are employed to assess different aspects of distributional similarity, including overall shape, central tendency, and tail behavior.

\subsubsection{Kolmogorov-Smirnov (KS) Test}
KS test is a non-parametric statistical test \cite{massey1951kolmogorov} that compares the empirical CDFs of two samples to determine whether they differ significantly. It is particularly effective in assessing how closely synthetic data replicates the distributional characteristics of real data, including behaviors in the extreme regions.
Given two samples of sizes $n$ and $m$, with corresponding empirical CDFs $F_n(x)$ and $G_m(x)$, the KS statistic is defined as follows:
\begin{equation}
D_{n,m} = \sup_x|F_n(x) - G_m(x)|
\end{equation}
where $\sup_x$ is the supremum over all values of $x$, $F_n(x)$ and $G_m(x)$ represent the proportion of real and synthetic data less than or equal to $x$, respectively. Intuitively, the KS statistic captures the maximum vertical distance between the two cumulative curves, offering a single and interpretable measure of distributional discrepancy. Lower KS values indicate a closer alignment between real and synthetic distributions, reflecting better preservation of statistical properties. 

The KS test is well-suited for extreme data generation due to its sensitivity to differences in both location and shape of the distributions, allowing the detection of large discrepancies in the critical tail regions. Moreover, its distribution-free nature enables broad applicability across domains without specific parametric assumptions.
In practice, it has been widely adopted to evaluate synthetic extreme data. For instance, it is applied to assess the fidelity of heavy-tailed financial distributions in Pareto-GAN \cite{huster2021pareto} and to quantify deviations in extreme precipitation patterns in GP-GAN \cite{gpgan}. It is further extended to sum the KS statistics across each dimension to handle multi-dimensional energy data in GMM-GAN \cite{zhao2024gan}. 

\subsubsection{The \textit{t}-test}
The \textit{t}-test is a parametric statistical test used to determine whether there is a statistically significant difference between the means of two samples \cite{kim2015t}. Unlike non-parametric tests for the entire distribution, the \textit{t}-statistic focuses specifically on central tendency, making it useful for verifying whether a generative model accurately captures the average behavior of real-world extreme data.
Let $\bar{x}$ and $s^2$ denote the mean and variance of the real sample data $x$ with size $n$, and $\bar{\tilde{x}}$ and $\tilde{s}^2$ represent the corresponding statistics for the synthetic sample $\tilde{x}$ with size $m$. 
The null hypothesis $H_0$ states that the mean of synthetic data is equal to that of the real data. The alternative hypothesis $H_1$ is that the means differ significantly.
The test statistic is calculated as:
\begin{equation}
t = \frac{\tilde{x} - x}{ \sqrt{\frac{s^2}{n} + \frac{\tilde{s}^2}{m}}}
\end{equation}
It quantifies how far the difference between two sample means deviates from zero, in units of the standard error. This statistic is compared with the critical value of the $t$ distribution, with degrees of freedom approximated using a variance-weighted formula. If the absolute value of \textit{t}-statistic exceeds this threshold, the null hypothesis is rejected, indicating a statistically significant difference between the synthetic and real means.

While the \textit{t}-test does not assess the full shape of the distribution, it provides a focused evaluation of whether the generative model captures the correct average behavior of extreme events. This is particularly valuable for extreme data applications where the average intensity, frequency, or duration carries operational significance.
For example, MC-TSGAN \cite{karimanzira2024mass} employed \textit{t}-test to validate that the mean of the generated flood discharge time series aligns with historical hydrological records, ensuring accurate preservation of the central tendency of extreme event modeling.

\subsubsection{Kullback-Leibler (KL) Divergence}

KL divergence is an information-theoretic measure \cite{castillo1997fitting} that quantifies how one probability distribution diverges from a reference probability distribution. For synthetic data generation, it captures information loss when using a synthetic distribution to approximate the real data distribution.
For two distributions $P$ and $Q$, with probability mass function $p(x)$ and $q(x)$ in the discrete case or density functions in the continuous case, the KL divergence from $Q$ to $P$ is defined as:

\begin{equation}
D_{\text{KL}}(P \parallel Q) = \mathbb{E}_{x \sim P} \left[\log \frac{p(x)}{q(x)}\right]
\end{equation}
where $\mathbb{E}_{x \sim P}$ denotes the expectation taken with respect to the distribution $P$. It is asymmetric, i.e., $D_{\text{KL}}(P \parallel Q) \neq D_{\text{KL}}(Q \parallel P)$. It also heavily penalizes scenarios where the synthetic distribution $q(x)$ assigns low or zero probability to regions where the real distribution $p(x)$ has substantial mass. Lower values indicate closer alignment, with zero implying a perfect match.

When applied to synthetic extreme data, KL divergence provides a fine-grained view of how well local density characteristics are captured, if the synthetic distribution sufficiently overlaps the real data. However, it can be unstable when the synthetic model fails to assign probability to tail regions in the real data, making it less robust than transport-based measures like the Wasserstein distance. As a result, it is often used in conjunction with other complementary metrics to more reliably assess the distributions.

In VAE-based modeling, KL divergence plays dual roles in both training and evaluation. For example, ExtVAE \cite{lafon2023vae} incorporated KL divergence into the ELBO objective to regularize the latent space, ensuring the alignment with heavy-tailed priors. Beyond training, a thresholded variant of KL divergence is applied as a performance metric to focus specifically on discrepancies in the tail regions, providing a targeted evaluation of how well rare but impactful events are captured.

\subsubsection{Wasserstein Distance}\label{sec:wassertein_distance}

The Wasserstein Distance, also known as Earth Mover's Distance \cite{arjovsky2017wasserstein}, measures the minimum cost of transforming one probability distribution into another. Conceptually, it interprets the distributions as piles of probability mass and calculates the total cost of moving this mass, where cost is defined as the amount of mass moved times the distance traveled.
In the univariate case, the first-order Wasserstein distance can be computed using the inverse CDFs (quantile functions) of two distributions:
\begin{equation} 
W = \int_0^1 \left| F^{-1}(u) - G^{-1}(u) \right| du 
\end{equation}
Here, $F^{-1}(u)$ and $G^{-1}(u)$ denote the quantile functions of the real and synthetic datasets, respectively, returning the value below which a proportion $u$ of the data falls. Empirically, this corresponds to sorting both datasets and computing the average distance between their corresponding quantiles.

Wasserstein distance is particularly well-suited for extreme scenarios, as it directly compares quantiles rather than relying on probability densities, and it remains robust even when the distributions have disjoint support. It captures not only central mismatches but also differences in location and spread of extreme values. 
To better focus on tail regions, recent studies \cite{yi2024method} proposed rescaled variants of Wasserstein distance, where only samples exceeding a specified threshold are considered and the resulting distance is normalized by the square of the threshold. This adjustment amplifies sensitivity to discrepancies among extreme samples \cite{lafon2023vae}. 

To address computational challenges in high-dimensional settings, the Sliced Wasserstein Distance emerges. It projects high-dimensional distributions onto multiple random one-dimensional directions and calculates the average Wasserstein distance between each of these projections. 
The tail-adapted variant has been further proposed \cite {girard2024deep}. which calculates the Euclidean norm of all data points, selects points exceeding high quantile thresholds (e.g., 90\%, 95\%, 99\%), normalizes them onto the unit sphere, and computes the distance between generated and real samples. This enables targeted and scalable evaluation of extreme-value behavior in high-dimensional data.

\subsubsection{Domain-Adapted Fréchet Inception Distance (FID)}
FID \cite{heusel2017gans} quantifies the distance between the two distributions in a latent feature space extracted from a deep convolutional network (typically the Inception-v3 model \cite{szegedy2016rethinking}), often evaluating the quality of synthetic images. It captures semantic similarity by comparing statistical moments (mean and covariance) of features, making it more robust to visual distortions and sensitive to the overall image structure. Formally, it is computed as:

\begin{equation}
    FID = ||\mu_r - \mu_g||^2 + Tr(\Sigma_r + \Sigma_g - 2(\Sigma_r\Sigma_g)^{1/2})
\end{equation}
where $(\mu_r, \Sigma_r)$ and $(\mu_g, \Sigma_g)$ represent the mean and covariance of the feature vectors for the real and generated samples, respectively, and $Tr$ is the trace of a matrix. Lower FID scores indicate that generated samples more closely match the distribution of real samples in feature space.

While FID is widely used in natural image synthesis, its direct application to synthetic extreme data can be problematic, as such samples often come from artificially generated image, such as weather maps or hazard simulations, rather than everyday visual scenes. The Inception-v3 network, pretrained on ImageNet, may fail to extract meaningful features, leading to unreliable or uninformative FID scores. To address this, ExGAN \cite{exgan} proposed a domain-adapted FID computation by training an autoencoder on target data and computing FID over bottleneck activations. This adaptation ensures that extracted features are semantically aligned with extreme target data, resulting in more accurate and sensitive evaluations of sample quality in tail regions \cite{gpgan}.

\subsection{Dependence Preservation}\label{sec:6.2}
Beyond matching marginal distributions, it is crucial for synthetic extreme data to preserve the dependence structures observed in real data. Metrics in this category capture multivariate dependencies, such as rank-based monotonicity and linear association, as well as temporal autocorrelation under extreme conditions.

\subsubsection{Absolute Kendall Error (AKE)}
AKE is a rank-based metric that evaluates how well synthetic data preserves the dependence structure of real multivariate data \cite{abdi2007kendall}. Rather than comparing variables individually, AKE captures how variables co-move, which is especially critical for modeling joint extreme events.
AKE is derived from Kendall's dependence functions, which describe the concordance structure between variable pairs.
Let $Z_{1,n} \leq ... \leq Z_{n,n}$ and $\tilde{Z}_{1,n} \leq ... \leq \tilde{Z}_{n,n}$ denote the sorted values from real and synthetic data, respectively, AKE is defined as:
\begin{equation}
AKE = \frac{1}{n} \times \sum |Z_{i,n} - \tilde{Z}_{i,n}|
\end{equation}
It corresponds to the 1-Wasserstein distance between the Kendall's dependence functions of two datasets. Lower AKE values indicate better preservation of the underlying dependency structure.

Since AKE compares ranks, it is robust to outliers and invariant to marginal transformations, making it effective for evaluating tail dependencies and joint behavior of rare event modeling.
EV-GAN \cite{allouche2022ev} used AKE to assess how well the synthetic data captures multivariate dependencies in extreme regions. 
HTGAN \cite{girard2024deep} similarly employed AKE with Sliced Wasserstein Distance to evaluate tail dependencies.

\subsubsection{Pearson Correlation}
The Pearson correlation coefficient  \cite{cohen2009pearson} is a statistical measure that quantifies the strength and direction of a linear relationship between two variables. Given two variables $X$ and $Y$ with standard deviations $\sigma_X$ and $\sigma_Y$, the Pearson correlation coefficient is defined as their covariance divided by the product of their standard deviations. Its value is between -1 and 1, where 1 implies perfect positive linear correlation, -1 implies perfect negative linear correlation, and $0$ indicates no linear correlation.

In synthetic extreme data generation, the Pearson correlation coefficient is often used to assess whether the linear dependencies between multiple variables are faithfully preserved. Misrepresenting correlations can significantly affect downstream tasks such as joint risk assessment, system planning, and extreme event forecasting. Several studies have employed Pearson correlation. InfoGAN \cite{yi2024method} measured the interdependencies among wind, solar, and load outputs in generated scenarios. Similarly, C-MCDCGAN-GP \cite{hongxia2023improved} compared the Pearson correlations of wind and PV outputs, showing better alignment with real extreme scenarios compared to baselines.

\subsubsection{Autocorrelation Functionc (ACF)}

ACF measures temporal dependence, the correlation between a time series and its lagged values \cite{cryer2008time}. It helps reveal important patterns, such as seasonality, trends, and random behavior, and supports statistical significance testing through confidence intervals.
Given a time series $\{x_t\}_{t=1}^{T}$, the lag-$k$ autocorrelation is:
\begin{equation} 
\rho(k) = \frac{\mathbb{E}[(x_t - \mu)(x_{t-k} - \mu)]}{\sigma^2} 
\end{equation} 
where $\mu$ is the mean and $\sigma^2$ is the variance of the series. The ACF curve plots $\rho(k)$ over a range of lags $k$, capturing the memory structure and smoothness of the time series. 

In synthetic extreme data generation, ACF is used to assess whether the temporal dynamics of extreme occurrences, such as bursts, clusters, or decay patterns, are preserved.  
A close match between ACF curves of real and synthetic data indicates that successful replication of sequential dynamics, while significant deviations suggest loss of temporal structure.
It is essential for applications such as renewable energy forecasting (e.g., solar, wind, and PV power), financial risk modeling, and environmental hazard prediction, where temporal dependencies significantly impact system dynamics under extreme conditions. 
For example, C-MCDCGAN-GP \cite{hongxia2023improved} applied ACF to evaluate the temporal relevance of synthetic wind and PV power data in renewable energy extreme scenarios.
Similarly, InfoGAN \cite{yi2024method} analyzed ACF curves across multiple lag intervals to assess whether synthetic wind-solar-load scenarios reflect the temporal correlation patterns in historical data. Besides, Tail-GAN \cite{tailgan} employed ACF up to 10 lags to validate that the synthetic price reproduces the structure of real market data.

\vspace{-0.1cm}
\subsection{Extreme Coverage}\label{sec:6.3}

It is crucial to ensure that extreme events themselves are sufficiently represented in synthetic data. Extreme coverage focuses on how well synthetic data captures the occurrence and variability of rare, high-impact scenarios.
Coverage rate measures the percentage of generated scenarios that satisfy a predefined extreme condition. A higher coverage rate indicates that synthetic data adequately reflects the variability and range of real extreme events. For example, in wind forecasting, the wind speeds above the 99th percentile (e.g., 25m/s) can be defined as extreme. If 8 of 100 generated samples satisfy it, then the coverage rate is 8\%, suggesting that the model underrepresents extreme occurrences.
Interval width measures the average spread between the upper and lower bounds of generated scenes, reflecting synthetic data's uncertainty. Continuing the example, for wind speed above 25 m/s, a model may produce ranges like 25–27 m/s or 25–35 m/s. While narrower intervals suggest higher confidence, they are not always better. Interval widths that closely align with real-world extreme intervals are preferred. 

Together, Coverage rate and Interval Width form a complementary evaluation framework: high coverage ensures that critical extremes are captured, while appropriately calibrated interval width avoids over- or underestimating uncertainty. Balancing these two aspects is essential for producing synthetic datasets that are both comprehensive and operationally useful. For instance, they were jointly used to assess the quality of synthetic wind–solar–load extreme scenarios \cite{yi2024method}, where the coverage rate evaluated the extent to which extreme patterns are captured, and the average width of power interval reflected the precision of generated scenarios. In addition, the coverage rate was also used in GMM-GAN \cite{zhao2024gan} to demonstrate the extremeness of synthetic wind power data.

\subsection{Extremal Dependence}\label{sec:6.4}
While Section~\ref{sec:6.2} addresses overall dependence across the full distribution, this section focuses specifically on dependence in the tails. Extremal dependence metrics quantify how strongly two variables behave together when rare and high-magnitude events occur.
The extremal coefficient~\cite{davison2012statistical} and extremal correlation coefficient~\cite{smith1990max} originate from EVT and are particularly suited for validating synthetic extreme data, where accurate modeling of joint rare occurrences is critical.
The extremal coefficient $\theta\in[1,2]$ measures the degree of upper-tail dependence between two variables:

\begin{equation}
    P(X \leq z, Y \leq z) = P(X \leq z)^\theta
\end{equation}
where $X$ and $Y$ are two random variables and $z$ is a high threshold. A value of $\theta=1$ indicates perfect dependence, meaning that if one variable exceeds a high threshold, the other does so almost surely. A value of $\theta=2$  indicates independence between extremes. Intermediate values correspond to partial dependence.

The extremal correlation coefficient $ \chi \in [0,1] $  provides an alternative view, quantifying the probability that one variable is extreme given that the other is extreme:
\begin{equation}
\chi = \lim_{u \to 1^{-}} \mathbb{P}\left(Y > F_Y^{-1}(u) \mid X > F_X^{-1}(u)\right)
\end{equation}
where $F_X^{-1}(\cdot)$ and $ F_Y^{-1}(\cdot)$ are the quantile functions of $X$  and $Y$ . A value of $\chi=1$ implies perfect co-occurrence of extremes, whereas $\chi=0$ indicates that extremes occur independently. The two metrics are mathematically related by:
$\chi=2-\theta$. Thus, while the extremal coefficient expresses the likelihood of simultaneous exceedance indirectly, the extremal correlation coefficient expresses it directly as a conditional probability. Both are essential for evaluating whether synthetic data accurately preserves the joint structure of extremes.

Capturing joint tail behavior is important to avoid underestimating compound risks in extreme data applications. In practice, Peard and Hall~\cite{peard2023combining} constructed extremal coefficient matrices to show that the GAN effectively replicated the extremal dependence for wind speed and wave height, although slightly underestimating negative dependence in precipitation, particularly over land areas. Similarly, EVT-GAN~\cite{boulaguiem2022modeling} computed extremal correlation coefficients across geographic locations, demonstrating that the joint occurrence probabilities of extreme precipitation events are preserved in synthetic spatial climate data. These highlight the importance of explicitly evaluating extremal dependence in synthetic data generation.

\subsection{Instance-Wise Reconstruction Loss}
\label{sec:6.5}
Reconstruction loss serves as a critical metric for evaluating how accurately extreme data generation models can reproduce unseen extreme samples \cite{zhu2016generative}. It directly quantifies instance-wise fidelity between original and generated images, typically using $L_1$ or $L_2$ norms. 
Given a real data point $x_i$ from a dataset with size $n$, where generation may be conditioned on auxiliary information $E(x_i)$ (e.g., severity level or latent embedding), and $G(z_i)$ is the generated sample from latent space vector $z_i$, the reconstruction loss is:
\begin{equation}
    L = \frac{1}{n}\sum_{i=1}^{n}\min_{z_i}||G(z_i, E(x_i)) - x_i||^2_2
\end{equation}
Lower loss values indicate that the generator preserves more of the fine-grained structure and key attributes. This instance-level evaluation complements metrics by revealing whether the model faithfully reconstructs individual rare events, not just approximates the distribution as a whole.

In practice, reconstruction loss has been incorporated alongside FID to improve evaluation of generation quality for extreme precipitation images \cite{exgan}. It demonstrates that minimizing the reconstruction error in extreme samples helps ensure that the generator learns to reproduce the fine-grained structural features that characterize extremes. This is especially important in domains where small deviations in structure, such as shape, intensity, or location of a weather cell, can have substantial semantic implications.

\subsection{Extreme Magnitude Accuracy}\label{sec:6.6}

These metrics evaluate generation accuracy through different perspectives: MAPE for relative error assessment, RMSE for absolute deviation measurement, and MSLE for tail-specific evaluation in heavy-tailed distributions, collectively ensuring that synthetic extreme data maintains both statistical fidelity and practical utility.
\subsubsection{Mean Absolute Percentage Error (MAPE)}
MAPE is a relative error metric that quantifies the average percentage difference between generated and actual values. By normalizing the error by the actual value, MAPE provides a scale-invariant measure of accuracy, making it particularly interpretable across different magnitudes of extremes. The general formula is:
\begin{equation}
MAPE = \frac{100\%}{n} \sum_{i=1}^{n} \left| \frac{x_i - \tilde{x}_i}{x_i} \right|
\end{equation}
where $x_i$ represents actual values and $\tilde{x}_i$ represents generated values. 

In extreme data generation, MAPE is widely used to assess whether synthetic outputs meet specific conditioning targets. ExGAN \cite{exgan} applied MAPE to assess how accurately generated samples achieve requested extremeness levels drawn from a Generalized Pareto Distribution. Tail-GAN \cite{tailgan} employed MAPE to compare VaR and ES for strategies computed under synthetic and real data, demonstrating that the model can capture dynamic market behaviors with stable performance.
Similarly, InfoGAN \cite{yi2024method} used MAPE to evaluate scenario-based renewable energy generation, comparing real and generated power outputs over time. In all cases, lower MAPE values indicate better performance, with values closer to zero representing a more accurate generation of extreme scenarios according to the target conditions.

\subsubsection{Root Mean Square Error (RMSE)}

RMSE measures the average magnitude of difference between produced and observed values, with larger deviations penalized more heavily  due to the squaring operation: 
\begin{equation}
RMSE = \sqrt{\frac{1}{n}\sum_{i=1}^{n}(x_i - \tilde{x}_i)^2}
\end{equation}

In extreme data generation, it is applied to assess the absolute deviation between generated and real curves, such as the power data in renewable energy forecasting \cite{yi2024method}. The goal is to ensure that generated scenarios not only replicate the general shape but also accurately match the magnitude of rare events, such as sudden power ramps or low-generation periods.
Compared to MAPE, RMSE places greater emphasis on larger errors, making it particularly suitable for identifying mismatches in extreme samples. Lower RMSE indicates that generative models produce outputs closer to the real data in absolute terms, which is critical when rare but impactful deviations must be accurately captured.

\subsubsection{Mean Squared Logarithmic Error (MSLE)}
MSLE is a scale-aware evaluation metric that measures the squared difference between the logarithms of predicted and actual values. It is particularly suited for data with heavy-tailed distributions or wide value ranges, as it compresses larger values and emphasizes relative rather than absolute accuracy. Therefore, it is particularly appropriate when extreme values vary across several orders of magnitude, such as in financial returns or hydrological extremes. The general formula is:
\begin{equation} 
MSLE = \frac{1}{n} \sum_{i=1}^{n} \left( \log(1 + x_i) - \log(1 + \tilde{x}_i) \right)^2 
\end{equation} 

For extreme data generation, evaluation often focuses on the tail of the distribution—the rare and high-magnitude samples that represent the most critical events. To specifically assess tail behavior, MSLE is adapted by restricting attention to the top $(1-\xi)\cdot n$ extreme samples above a certain quantile threshold $\xi\in\{0.90,0.95,0.99\}$: 
\begin{equation} 
MSLE(\xi) = \frac{1}{d \cdot m} \sum_{j=1}^{d} \sum_{i=1}^{m} \left( \log x_i^{(j)} - \log \tilde{x}_i^{(j)} \right)^2 
\end{equation} 
where $d$ is the dimensionality, $m=(1-\xi)\cdot n$ is the number of tail samples per dimension, and $x_i^{(j)}$, $\tilde{x}_i^{(j)}$ denote the real and generated tail samples in dimension $j$. 

To further enhance sensitivity to multiple dimensions, MSLE is refined in EV-GAN \cite{allouche2022ev}. Instead of comparing randomly selected tail samples, it computes MSLE over sorted order statistics, aligning the $i^{th}$ largest real value with the $i^{th}$ largest generated value in each dimension. Therefore, the real and generated samples $x_i^{(j)}$ and $\tilde{x}_i^{(j)}$  are replaced by their respective order statistics $x_{n_i+1,n}^{(j)} $ and $\tilde{x}_{n_i+1,n}^{(j)}$. This modification allows direct comparison of the shape and scale of tail distributions, ensuring that the model accurately reproduces the relative intensity of extreme values. Lower MSLE values indicate stronger performance in capturing the tail behaviors, particularly sensitive to the discrepancies in relative scale among the most extreme samples.
\subsection{Visualization Diagnostics}\label{sec:6.7}
Beyond quantitative metrics, visualization techniques provide complementary intuitive insights. We distinguish between dimensionality reduction techniques (PCA and t-SNE) that project high-dimensional data into interpretable visual spaces, and empirical distribution visualization methods (Q-Q plots and rank-frequency distributions) that directly compare distributional characteristics and extreme behaviors. 
\subsubsection{Principal Component Analysis (PCA)}
PCA is a dimensionality reduction technique that identifies the directions of greatest variance within the data \cite{mackiewicz1993principal}. It constructs a new set of orthogonal axes, known as principal components, which are linear combinations of the original features and ordered by explained variance. The first principal component captures the maximum variance, the second captures the largest remaining variance orthogonal to the first, and so forth.
The process begins by centering the data by its sample mean and computing the eigenvectors and eigenvalues of its covariance matrix. Each eigenvector defines the direction of a principal component, and the corresponding eigenvalue quantifies the variance explained along that direction. By selecting the top-$k$ eigenvectors associated with the largest eigenvalues, the data can be projected into a $k$-dimensional subspace, preserving as much of the original variability as possible. PCA plot thus visualizes complex and high-dimensional datasets in two or three dimensions.

In the context of synthetic extreme data generation, PCA serves as an intuitive visual diagnostic tool to compare real and synthetic distributions. By projecting both datasets into the same lower-dimensional principal component space, researchers can assess whether the synthetic data replicates the dominant variability structure of the real data and extends into rare, high-magnitude regions that are critical for extreme modeling. It has been applied to evaluate synthetic flood event generation, identifying the  principal components explaining the most variance in flood data and enabling direct visual comparison between real and synthetic samples \cite{karimanzira2024mass}.

\subsubsection{t-Distributed Stochastic Neighbor Embedding (t-SNE)}
t-SNE is a non-linear dimensionality reduction method to visualize high-dimensional data \cite{van2008visualizing}. Unlike linear methods like PCA, which focus on maintaining global structure and maximum variance, t-SNE preserves local neighborhood relationships. Points that are close in the high-dimensional space are mapped close together in the low-dimensional embedding, while distant points are placed as far apart as possible.
The process first converts pairwise Euclidean distances between points into conditional probabilities that represent similarities. modeled with a Gaussian distribution in the high-dimensional space. In the low-dimensional space, similarities are modeled using a Student’s t-distribution to mitigate the crowding problem that can arise when reducing many dimensions to two or three. Then it minimizes the KL divergence between these two distributions, iteratively optimizing the low-dimensional embedding to best preserve local structures.

For extreme data, t-SNE visualizes whether synthetic data captures fine-grained structures and neighborhood patterns of the real data. By comparing t-SNE plots of real and synthetic datasets, researchers can evaluate whether synthetic samples preserve local clusters and complex manifolds. This is particularly important in extreme data where subtle variations at tails or rare event clusters may carry critical information.
In practice, t-SNE has been used alongside PCA in evaluating synthetic flood event generation, highlighting differences and overlaps in local structures between real and generated data \cite{karimanzira2024mass}. Together, these visualizations offer a complementary view of distributional fidelity at both global and local scales.

\subsubsection{Quantile-Quantile (Q-Q) Plots}

The Q-Q plot is a graphical tool to compare the distribution of two datasets or to compare empirical data against a theoretical reference distribution, providing an intuitive visual assessment of distributional similarity \cite{marden2004positions}.
To construct a Q-Q plot, samples from each dataset are sorted in ascending order, and their corresponding quantile values are plotted against each other. If both datasets share the same underlying distributions, the points should approximately align along the $y=x$ diagonal line. Deviations from this line indicate differences in distributional properties, such as location, scale, skewness, or tail behavior. 

To better emphasize tail differences, log-transformed Q-Q plots (log-Q-Q plots) are often employed. By plotting the logarithm of the quantiles instead of raw values, these plots stretch the extreme regions and compress the center, making deviations in the tails more visible. A linear relationship in the log-Q-Q plot, particularly at high quantiles, suggests accurate modeling of heavy-tailed or power-law behaviors.

Both standard and log Q-Q plots have been widely adopted in evaluating synthetic extreme data quality. For instance, in climate modeling, Q-Q plots were used to compare the distribution of the extremal coefficient across training, test, and generated datasets for wind speed data across all possible pixel pairs \cite{peard2023combining}. Their analysis revealed that while the model captured general extremal structures, it tended to underestimate dependence between the most extreme events, an important finding for disaster risk assessment.
Similarly, log Q-Q plots were in VAE-based models \cite{lafon2023vae} to demonstrate the accurate modeling of heavy-tailed behavior, where synthetic samples closely followed the expected linear patterns in the tails. This visualization complements numerical metrics and intuitively validates the realism of generated extreme values.

\subsubsection{Rank-Frequency Distribution}

The rank-frequency distribution is a statistical visualization method that represents the relationship between the frequency or magnitude of an observation and its rank \cite{hill1974rank}. 
To construct a rank-frequency distribution, data points are ranked in descending order by their frequencies, with rank plotted on the x-axis and frequency on the y-axis. For continuous variables, discretization or binning is often applied prior to ranking.  

This approach is particularly useful for visualizing the distribution of extreme values or outliers by matching the shape and decay patterns between the real and synthetic datasets.
To better highlight tail behavior, axes are plotted on logarithmic scales. If the underlying distribution follows a power-law decay, the plot on a log-log scale will exhibit an approximately linear trend, especially in the extreme regions. Deviations from linearity can reveal underestimation or overestimation, which is critical in risk-sensitive applications.
For example, Tail-GAN \cite{tailgan} utilized rank-frequency distribution to evaluate how accurately synthetic financial data captures extreme risk events across various trading strategies. By plotting the rank-frequency distributions of strategy profits and losses on logarithmic scales, they visually compare the tail performance of real market data against synthetic scenarios. This visual comparison provides direct evidence of the model's ability to reproduce lower-tail quantiles associated with significant financial profit and losses.

\subsection{Downstream Performance Validation}\label{sec:6.8}
Unlike earlier sections that focus on direct evaluation—comparing synthetic and real data directly—this section introduces a unique and complementary perspective: indirect evaluation through downstream tasks. In this approach, synthetic data is incorporated into a real-world training pipeline, and its utility is assessed by measuring the performance improvements in downstream prediction or decision-making tasks. This type of evaluation tests not only the visual or statistical fidelity of synthetic data but also its practical value in enhancing model generalization and robustness under extreme conditions.
The typical process involves augmenting real datasets with synthetic extreme scenarios, retraining downstream models, and comparing predictive performance before and after the integration of synthetic data. Differences in results suggest whether the synthetic data captures meaningful extreme patterns and contributes to better generalization and risk forecasting.
For example, Jiang et al. \cite{jiang2024leveraging} treated high-risk financial market events as extremes and focused on forecasting high-risk event occurrence. They evaluated downstream performance by comparing classification accuracy and F1 score before and after adding synthetic samples. Results showed that introducing synthetic extreme data improved the downstream model’s ability to detect market irregularities, demonstrating the practical value of synthetic augmentation.
Similarly, Karimanzira \cite{karimanzira2024mass} used synthetic flood event data to improve probabilistic flood-forecasting models for predicting the discharge confidence interval. Widely used hydrological metrics, such as Nash-Sutcliffe Efficiency \cite{waseem2017review} and Kling-Gupta Efficiency \cite{waseem2017review}, validated that the model better captured the observed rainfall–runoff relationship. Interval-based metrics, such as Mean Prediction Interval Width and Prediction Interval Coverage Probability, as mentioned in Section \ref{sec:6.3}, evaluate the precision and reliability of prediction intervals. They also reflect improved forecasting performance after integrating synthetic extreme data. 
Overall, indirect evaluation complements direct evaluation by validating not just the visual or statistical similarity of synthetic data, but also its effectiveness in improving real-world predictive tasks under extreme conditions.

\section{Domain-Specific Applications}\label{sec:domain}

\begin{table}
\caption{Domain-specific tasks in synthetic extreme data generation}\centering\small
\begin{tabular}{m{2.5cm} m{5.5cm} m{4cm}}
\toprule
\textbf{Domain} & \textbf{Extremes / Tasks} & \textbf{Future Directions} \\
\midrule
\multirow{3}{*}{\centering Finance}
& Market crashes \cite{tailgan, chen2024risk, huster2021pareto} & Behavioral anomalies \\
& Fraud and financial misconduct\cite{jiang2024leveraging, azimi2024zgan} & Macroeconomic policy shock\\
& & DeFi-related disruptions\\
\midrule
Healthcare & Rare disease modeling \cite{disease} & Wide-range infectious outbreaks\\
\midrule
\multirow{3}{*}{\centering Climate}
& Precipitation/temperature extremes \cite{exgan, boulaguiem2022modeling} & Wildfire event generation \\
& Climate emulator refinement \cite{puchko2020deepclimgan, bassetti2024diffesm} & Windstorm hazard modeling \\
& Coherent multivariate weather events \cite{peard2023combining, klemmer2021generative} & \\
\midrule
Renewable Energy & Wind-solar-load extremes \cite{yi2024method, li2023c, hongxia2023improved} & \\
\midrule
Hydrology & Flood discharge simulation \cite{karimanzira2024mass} & \\
\midrule
Geophysics & & Earthquake hazard simulation\\
\bottomrule
\end{tabular}
\label{tab:domain_tasks}
\end{table}

Synthetic extreme data generation plays a crucial role across domains where rare, high-impact events pose major risks, including finance, healthcare, climate, renewable energy, hydrology, and geophysics. Each domain presents unique modeling challenges: finance involves human-driven behavioral dynamics, healthcare faces critical data scarcity for rare diseases and pandemic preparedness, climate requires capturing complex spatiotemporal patterns, renewable energy must model stochastic intermittency under extreme conditions, hydrology integrates physical constraints with statistical modeling, and geophysics represents an emerging frontier for seismic risk assessment. This section reviews key application areas, highlights recent advancements, and outlines emerging directions, summarized in Table \ref{tab:domain_tasks}.

\subsection{Finance}

The finance domain encompasses both established and emerging extreme event applications for synthetic data generation, where human-driven dynamics create unique modeling challenges distinct from natural disasters governed by physical laws. Given that financial extremes are shaped by behavioral psychology, socio-economic structures, and collective decision-making—as exemplified by the 2008 subprime crisis—synthetic data offers a powerful solution to overcome the scarcity and limited diversity of historical extreme event data. Current research has successfully addressed market crashes through tail risk modeling and market supervision via rare event detection for fraud and manipulation. However, as shown in Table 3, three critical underexplored areas marked with asterisks present significant opportunities: behavioral anomalies driven by collective investor psychology (exemplified by events like the GameStop short squeeze), macroeconomic policy shocks from unprecedented fiscal interventions (such as COVID-19-era policies), and DeFi-related disruptions involving protocol exploits and smart contract vulnerabilities. These emerging domains require sophisticated synthetic approaches to model the complex interplay of psychological dynamics, policy impacts, and cascading failures characteristic of modern decentralized financial systems.

\subsubsection{Human-Driven Nature of Financial Extremes}
Extreme financial scenarios, such as market crashes, liquidity crises, credit defaults, and systemic banking failures, can lead to cascading disruptions across global economies \cite{gu2023margin}. Due to their rarity and potential devastation, modeling these events is critical for robust risk management, stress testing, and regulatory compliance.

Unlike natural disasters, such as floods or earthquakes, which are governed by immutable physical laws, financial extremes are driven by human behavior, socio-economic structures, and collective decision-making. For instance, the 2008 subprime mortgage crisis was not a natural inevitability but a human-driven catastrophe fueled by speculative housing investments, excessive leverage, and insufficient regulatory oversight. Market psychology, including greed, fear, herd behavior, and rumor propagation, amplified systemic vulnerabilities, making financial extremes dynamic and difficult to predict.

This human-centric characteristic introduces both a challenge and an opportunity: while historical data on extreme financial events is scarce and often fails to capture sufficient diversity and severity, synthetic data generation offers a powerful alternative to simulate plausible future scenarios under varying behavioral and structural assumptions. Synthetic rare-event data allows institutions to train predictive models on enriched distributions with edge cases, supporting more robust and anticipatory decision-making.

\subsubsection{Market crashes}Several specific financial extremes have already been explored. For tail risk modeling, Tail-GAN \cite{tailgan} simulated scenarios in the extreme tails of loss distribution for dynamic portfolios, accurately preserving key financial risk characteristics such as Value-at-Risk (VaR) and Expected Shortfall (ES). This capability allowed financial institutions to conduct robust stress tests and assess the resilience of portfolio strategies under severe, hypothetical market shocks that are not fully represented in past data. 
Similarly, Pareto GAN \cite{huster2021pareto} refined loss functions to stabilize training convergence on heavy-tailed financial distributions, making it better suited for extreme-risk financial applications such as credit risk estimation, systemic banking crises, and market stress testing. 
\subsubsection{Fraud and financial misconduct}
GAN-based models were also leveraged by financial regulatory agencies to generate synthetic data for rare event classes relevant to market supervision \cite{jiang2024leveraging}, such as fraud, market manipulation, or abrupt systemic risk. By augmenting rare-event samples, synthetic data improves the performance of risk detection frameworks and provides more informative data for regulatory monitoring, enhancing early-warning capabilities for financial supervision. Another important application lies in banking and insurance, where rare outcomes such as defaults, fraud, or customer churn are central concerns. zGAN \cite{azimi2024zgan} generated synthetic samples with outlier characteristics to balance highly imbalanced datasets, improving the predictive power and reliability of classification models used for fraud detection, churn prediction, and credit risk assessment.

\subsubsection{Behavioral anomalies}Despite these advances, several critical areas in finance remain underexplored. One promising direction is behavioral finance-driven events, where collective action, rumor propagation, speculative bubbles, and sentiment-driven trading can trigger abrupt and irrational market movements. A representative example is the 2021 GameStop short squeeze, where retail investors, largely organized through social media platforms like Reddit, collectively drove up the price of GameStop stock, forcing large institutional short-sellers into massive losses. Synthetic data generation can simulate behavioral patterns, such as feedback loops, panic selling, or coordinated speculation, to better anticipate the impact of investor psychology on market outcomes under varying conditions. 
\subsubsection{Macroeconomic policy shock}Another important area is macroeconomic policy shock, where abrupt fiscal or monetary interventions, such as sudden interest rate hikes or currency devaluations, can trigger rare but impactful regime changes in financial systems. A major instance is the COVID-19-era policy interventions, during which global central banks and governments deployed unprecedented policies, including zero or negative interest rates, massive quantitative easing, and direct stimulus payments. These interventions distorted traditional asset dynamics, liquidity flows, and risk-return profiles in ways not captured by historical data. Synthetic generation of extraordinary policy response scenarios enables institutions to stress-test market resilience, evaluate systemic vulnerabilities, and explore the ripple effects of extreme policy moves across asset classes and time horizons.
\subsubsection{DeFi-related disruptions}As an emerging and rapidly growing field, DeFi-related disruptions, characterized by protocol exploits, smart contract vulnerabilities, and on-chain liquidity crises in blockchain-based financial systems, represent a rapidly evolving frontier. A key example is the 2022 Wormhole bridge hack, where over \$300 million in crypto assets were stolen due to a smart contract vulnerability in a cross-chain protocol. Such events are difficult to model using historical data alone, highlighting the need to simulate adversarial attacks, liquidity shocks, governance failures, and cascading protocol interdependencies to understand system risk contagion and to design more resilient decentralized systems.

\subsection{Healthcare}
Healthcare extremes present critical modeling challenges where data scarcity impacts patient outcomes and public health response. Current research successfully addresses rare disease modeling, using generative AI for conditions like Huntington's disease while enabling privacy-preserving collaborative research. However, wide-range infectious outbreaks remain underexplored despite their critical importance, as demonstrated by H1N1, Ebola, and COVID-19 where incomplete real-time data limited effective response. Synthetic generation can simulate infection trajectories and healthcare scenarios, strengthening pandemic preparedness systems.
\subsubsection{Rare disease modeling}In the healthcare domain, synthetic data is increasingly used for rare disease research data, where the limited high-quality and representative datasets have been a long-standing barrier to the development and validation of robust machine learning models and clinical research efforts. Conditions such as Huntington’s disease, progeria, or certain pediatric cancers exemplify the challenges in collecting sufficient patient data. 
Recent work \cite{disease} highlighted that generative AI models could produce high-quality synthetic data, such as survey response data and electronic health records, to reflect diverse manifestations and progression patterns of rare diseases. This enables the development of better diagnostic tools and treatment strategies. Importantly, since synthetic data can be generated without direct patient identifiers, it provides a powerful means of privacy-preserving data sharing across institutions, which is crucial for collaborative rare disease research while remaining compliant with regulations like the Health Insurance Portability and Accountability Act (HIPAA) and the General Data Protection Regulation (GDPR).

\subsubsection{Wide-range infectious outbreaks}Additionally, wide-range infectious outbreaks, although not always rare in occurrence, often exhibit rare combinations of clinical, demographic, and geographic patterns, especially during early or extreme phases of pandemics. Despite its critical importance, this area remains underexplored in synthetic data generation research. Events such as the 2009 H1N1 influenza pandemic and the 2014-2016 Ebola outbreak underscore the global consequences of limited preparedness for extreme-scale infections. The early-stage spread of COVID-19 exhibited highly heterogeneous patient responses, overwhelmed hospital systems, and regionally varying resource constraints. In such crisis, real data is often incomplete, noisy, or delayed, severely limiting the effectiveness of real-time modeling and policy response. Synthetic data generation can help simulate various “what-if” infection trajectories, intervention outcomes, and healthcare burden scenarios under different assumptions regarding viral mutations, patient comorbidities, vaccination uptake, or policy decisions. Therefore, it can improve the resilience of public health forecasting systems, support stress testing of hospital infrastructure, and enable the training of AI models robust to extreme clinical surges. As future pandemics or biosecurity threats may present even more complex and extreme patterns, building a foundation of synthetic scenario generation in wide-scale infections is both a promising and urgently needed direction.

\subsection{Climate and Weather Extremes}
Severe atmospheric phenomena, such as floods, hurricanes, wildfires, and windstorms, are becoming increasingly frequent due to climate change, posing major risks to infrastructure, agriculture, and societal preparedness. Generating synthetic climate scenarios is valuable for risk assessments, disaster preparedness, and infrastructure planning. Several studies have adopted GAN-based models for extreme weather. 

\subsubsection{Precipitation and temperature extremes}Precipitation and temperature extremes have been a primary focus.
ExGAN \cite{exgan} and EVT-GAN \cite{boulaguiem2022modeling}  specifically targeted the tails of climate variable distributions and validated their models on U.S. and European precipitation records for realistic simulation on spatial climate risks. 

\subsubsection{Coherent multivariate weather events}Beyond univariate extremes, there was a growing emphasis on the synthesis of multivariate and spatially coherent events, reflecting the reality that extreme weather hazards (like flooding or cyclones) often result from the interaction of multiple meteorological variables (e.g., total precipitation, significant wave height, wind speed) over large geographical areas. For example, Peard and Hall \cite{peard2023combining} modeled the dependence structure between multiple climate hazards, such as simultaneous high winds, waves, and rainfall.

\subsubsection{Climate emulators}Climate emulators like DeepClimGAN \cite{puchko2020deepclimgan} and DiffESM \cite{bassetti2024diffesm} refined the temporal and spatial resolution of Earth System model outputs from monthly to daily scales to better capture short-duration extremes like floods.
Additionally, conditioning techniques, such as detected extreme weather event masks, have been incorporated into GANs \cite{klemmer2021generative} to explicitly guide the synthesis process, ensuring the realism and statistical fidelity of simulated climate hazards.
Evaluations typically incorporate both statistical similarity and domain-specific metrics, such as tail coverage and physical constraints, ensuring that synthetic climate scenarios are not only statistically plausible but also physically consistent, making them valuable for adaptation planning under increasing climate risk.

\subsubsection{Wildfire event generation}While significant progress has been made in modeling precipitation extremes and compound hazards, certain types of climate-driven extremes remain promising but still underexplored. In particular, {wildfire event,
exacerbated by prolonged droughts, rising temperatures, and changing land use—have caused catastrophic losses in regions like California, Australia, and Southern Europe, affecting not only ecosystems but also air quality, public health, and economic stability.

\subsubsection{Windstorm hazard}Meanwhile, windstorm hazard, including derechos and extratropical cyclones, has led to widespread infrastructure damage, disrupted energy systems, and cascading societal impacts. Simulating such extremes is essential for proactive risk assessment, resilience planning, and insurance stress testing.  Given the increasing societal, environmental, and economic impacts of wildfires and severe wind events, developing synthetic data frameworks for these phenomena should be prioritized for future work in climate extreme modeling.

\subsection{Renewable Energy}
The integration of renewable energy sources, such as wind and solar, into power grids introduces uncertainty due to the stochastic and intermittent nature of natural resources. Extreme weather conditions, such as prolonged droughts, high wind events, or simultaneous low-wind and low-sunlight conditions, can lead to sudden losses in energy output, requiring precise extreme-event data synthesis for risk-aware energy system management.

GAN-based models have been developed to simulate such extremes. InfoGAN \cite{yi2024method} focused on generating synthetic wind-solar-load extreme scenarios, capturing how weather, extreme meteorological conditions, and natural disasters affect renewable energy outputs. Conditions of no sunlight and no wind are regarded as typical extreme scenarios within power systems due to their broad impact and high occurrence frequency. Therefore, GAN architectures have been modified to capture the correlation dynamics between energy sources and generate synthetic wind-solar-load extreme scenarios. 
C-DCGAN \cite{li2023c} extended it by embedding uncertainty measures, providing tailored extreme synthetic datasets for energy system risk analysis.
In addition, multi-channel GANs  \cite{hongxia2023improved} generated high-resolution joint wind-PV energy scenarios. These improvements help refine energy storage optimization strategies and grid reliability forecasting.

These synthetic extreme scenarios support power grid stress testing, energy market forecasting, and operational planning under rare but impactful conditions, with many models incorporating physical constraints to ensure outputs remain plausible within known energy system dynamics.

\subsection{Hydrology}
Hydrological extremes, particularly floods and river discharge surges, represent critical risks to urban and agricultural planning. 
Unlike atmospheric phenomena, hydrological extremes result from complex surface and subsurface processes triggered by meteorological forcing, such as heavy precipitation or snowmelt. Thus, synthetic data generation for hydrological systems must capture not only the statistical properties of precipitation inputs but also the physical dynamics of water movement through landscapes and river networks.

MC-TSGAN \cite{karimanzira2024mass} introduced a GAN-based model that synthesized flood discharge scenarios while embedding hydrological principles into the loss, improving the model's applicability for risk assessment and disaster prevention. 
These synthetic scenarios improved probabilistic forecasting models on predicting flood discharge and water level prediction intervals under extreme rainfall conditions, supported multi-step-ahead runoff predictions, and informed stability analyses under various coincident extreme conditions. It enabled the power system to perform robust risk analysis, stress testing, and scenario-based operations planning for hydropower and urban water management. 

In hydrological modeling, evaluation criteria typically extend beyond conventional statistical distribution matching. It is essential to assess physical constraints relevant to the energy system, like mass conservation, energy balance, or hydrological principles, ensuring that synthetic data remains not just statistically plausible, but also consistent with domain knowledge.

\subsection{Geophysics}
Beyond atmospheric and hydrological extremes, geophysical hazards such as earthquakes represent another critical domain for synthetic extreme data generation. Earthquake-induced disasters, including ground shaking, surface rupture, tsunamis, and cascading infrastructure failures, have led to devastating societal and economic consequences, as seen in events like the 2010 Haiti earthquake and the 2011 Tohoku earthquake in Japan. Synthetic earthquake scenarios are valuable for seismic risk assessment, urban infrastructure design, insurance modeling, and emergency response planning.

While generative approaches for seismic event synthesis are less mature compared to climate and hydrology applications, research \cite{yamagishi2021spatio, munchmeyer2021earthquake,schiappapietra2020modelling} focuses on modeling earthquake magnitudes, ground motion fields, spatial-temporal event clustering, and correlated damage patterns across regions. Generating synthetic catalogs of seismic events allows for stress testing structural designs, evaluating the resilience of critical infrastructure networks (e.g., transportation, utilities, hospitals), and informing probabilistic seismic hazard assessments under rare but catastrophic conditions.

Given the limited frequency of large-scale real-world earthquake events and the inherent uncertainty in future seismic activity, expanding synthetic extreme data research into geophysical extremes represents a promising and urgently needed frontier. Reliable synthetic seismic datasets could substantially enhance preparedness strategies, support urban resilience planning, and improve loss estimation frameworks for high-risk regions worldwide.

\section{Challenges}\label{sec:challenge}
Despite significant advances in synthetic extreme data generation, several fundamental challenges persist. Data-related challenges include extreme-specific preprocessing demands and scarcity of high-quality event data, particularly for interconnected systems. Methodologically, a critical trade-off exists between generating realistic samples and diverse scenarios. Additionally, cross-domain consistency remains challenging when modeling cascading effects across interconnected systems. These challenges highlight key areas requiring continued research attention.

\subsection{Extreme-Specific Preprocessing Demands}
Preprocessing extreme event data poses significant challenges due to the need for domain expertise, ambiguous labeling for extremeness, and resolution mismatches. (1) Deep domain expertise is often required to resolve missing values, corrupted records, and inconsistent formats. It is particularly for domains like wildfires, earthquakes, and windstorms, where data originates from diverse sources, such as remote sensing, sensor networks, government logs, and crowdsourced reports, all varying in quality, completeness, and granularity. These steps are labor-intensive and often require manual tuning guided by expert judgment.
(2) Determining what qualifies as “extreme” is complex, as rare events are often ambiguously annotated (e.g., thresholds for “severe” damage vary) or labeled inconsistently across jurisdictions and data providers. This lack of standardization complicates the construction of reliable ground truth for supervised learning. 
(3) Temporal and spatial resolution mismatches are common: satellite wildfire imagery may be updated daily while associated weather data is recorded hourly, or windstorm damage data may be aggregated by region but misaligned with meteorological measurements on dense grids. Aligning such data demands resampling, interpolation, and careful coordination across spatial and time scales.

\subsection{Data Scarcity From Isolated Events to Interconnected Extremes}

A central challenge is the scarcity of high-quality, well-annotated data, not only for individual extreme events, but also for the auxiliary signals and their interdependencies critical for understanding and modeling such events.
(1) Public datasets are often fragmented across domains, institutions, or geographic regions, and many events of interest, such as blackouts, cyberattacks, or disease outbreaks, are underreported or inaccessible due to privacy, policy, or security constraints.
(2) It is exacerbated by the lack of auxiliary contextual signals, such as meteorological conditions, infrastructure states, or socioeconomic indicators, which are critical for capturing the dynamics and drivers of rare events. They are often missing, misaligned, or recorded at incompatible temporal or spatial resolutions. 
(3) Most importantly, modeling interdependent systems poses a unique challenge, as extreme events in one domain can trigger cascading effects in others. Cross-domain data to capture such dependencies is even harder to obtain than isolated-event data, as it requires sufficient, synchronized data across multiple systems that is rarely available.
This data gap limits the development of causal and conditional modeling and complicates validation and interpretability.

\subsection{Trade-off Between Realism and Diversity}

A fundamental tension in synthetic rare-event generation lies in balancing realism with diversity. Generative models that closely replicate historical patterns may overfit to the limited examples of observed extremes, failing to produce useful counterfactuals or simulate novel but plausible edge cases. Conversely, models optimized for diversity may generate scenarios that violate domain-specific constraints or physical laws, reducing their credibility and utility.
This trade-off is especially difficult in rare-event modeling, where data sparsity limits the range of legitimate variations the model can learn from. 
For this reason, Wasserstein distance in Section \ref{sec:wassertein_distance}, which captures overall distributional similarity including tail behavior, is often used to evaluate realism, and should be considered alongside coverage rate in Section \ref{sec:6.3}, which serves as a diversity metric by quantifying how well rare events are represented. 
Ensuring that generated data remains both realistic and sufficiently varied, while respecting structural, temporal, or causal dependencies, remains a key open challenge in high-risk domains.

\subsection{Cross-Domain Consistency}
Extreme events often propagate across interconnected systems. For example, during Hurricane Sandy, power grid failures disrupted water treatment plants, which in turn limited clean water access for hospitals and emergency shelters. Simultaneously, flooded transportation routes delayed the deployment of maintenance crews and emergency supplies. 
Models on isolated components, such as electricity outages without considering transportation or water system dependencies, would miss the compound nature of the crisis and lead to misleading conclusions in stress testing or disaster planning.
However, it is challenging, as generative models must preserve correlations, causal relationships, and shared constraints across heterogeneous data types and resolutions. 
Metrics such as Pearson correlation and AKE in Section \ref{sec:6.2} can assess inter-variable associations across domains, while extremal correlation coefficients evaluate tail dependencies in joint extreme events. 
Robust modeling of these interactions remains an open research problem, requiring structured conditioning, multi-modal learning, and domain-specific constraints.

\section{Conclusion}\label{sec:conclusion}
This survey provides the first comprehensive review of synthetic data generation for extreme events, moving beyond prior surveys that focus on privacy concerns.
We examine a wide range of techniques, benchmark datasets, and applications across critical domains. 
Our core contribution is a structured evaluation framework tailored to extremeness, covering diverse aspects and offering in-depth insights into metric applicability and domain-specific adaptations for rigorous assessment. We also identify high-impact yet underexplored application areas, such as behavioral finance, epidemic modeling, and natural disasters, and outline open challenges related to data interdependencies and modeling complexity. 
As the risks and complexities of extreme events grow, we hope this survey offers a foundation for future research and the development of resilient, risk-aware decision-making systems.  

\bibliographystyle{ACM-Reference-Format}
%%% -*-BibTeX-*-
%%% Do NOT edit. File created by BibTeX with style
%%% ACM-Reference-Format-Journals [18-Jan-2012].

\end{document}